\documentclass{article}
\usepackage{ijcai25}

\usepackage{times}
\usepackage{soul}
\usepackage{url}
\usepackage[hidelinks]{hyperref}
\usepackage[utf8]{inputenc}
\usepackage[small]{caption}
\usepackage{graphicx}
\usepackage{amsmath}
\usepackage{amsthm}
\usepackage{booktabs}
\usepackage{algorithm}
\usepackage[switch]{lineno}

\usepackage[T1]{fontenc}    % use 8-bit T1 fonts
\usepackage{amsfonts}       % blackboard math symbols
\usepackage{nicefrac}       % compact symbols for 1/2, etc.
\usepackage{microtype}      % microtypography
\usepackage{xcolor}         % colors
\usepackage{algpseudocode}
\usepackage{multirow}
\usepackage{enumitem}

\usepackage[normalem]{ulem}
\useunder{\uline}{\ul}{}

\newtheorem{definition}{Definition}

\newcommand{\Model}{OPEN}

\title{Towards Comprehensive and Prerequisite-Free Explainer for Graph Neural Networks}

\author{
Han Zhang$^1$
\and
Yan Wang$^1$\footnote{Corresponding author}
\and
Guanfeng Liu$^1$
\and
Pengfei Ding$^1$
\and 
Huaxiong Wang$^2$
\and 
Kwok-Yan Lam$^2$\\
\affiliations
$^1$Macquarie University\\
$^2$Nanyang Technological University\\
\emails
\{han.zhang13, pengfei.ding2\}@hdr.mq.edu.au,
\{yan.wang, guanfeng.liu\}@mq.edu.au,
\{hxwang, kwokyan.lam\}@ntu.edu.sg
}
\begin{document}

\maketitle

\begin{abstract}
    To enhance the reliability and credibility of graph neural networks (GNNs) and improve the transparency of their decision logic, a new field of explainability of GNNs (XGNN) has emerged. However, two major limitations severely degrade the performance and hinder the generalizability of existing XGNN methods: they (a) fail to capture the complete decision logic of GNNs across diverse distributions in the entire dataset's sample space, and (b) impose strict prerequisites on edge properties and GNN internal accessibility. To address these limitations, we propose \Model, a novel c\textbf{O}mprehensive and \textbf{P}rerequisite-free \textbf{E}xplainer for G\textbf{N}Ns. \Model, as the first work in the literature, can infer and partition the entire dataset's sample space into multiple environments, each containing graphs that follow a distinct distribution. \Model~further learns the decision logic of GNNs across different distributions by sampling subgraphs from each environment and analyzing their predictions, thus eliminating the need for strict prerequisites. Experimental results demonstrate that \Model~captures nearly complete decision logic of GNNs, outperforms state-of-the-art methods in fidelity while maintaining similar efficiency, and enhances robustness in real-world scenarios.
\end{abstract}

\section{Introduction}
Graph neural networks (GNNs), known for their capability to learn complex relational patterns in graphs, have gained significant attention and been widely applied in critical fields such as finance~\cite{rgnnsm_2022,xu2024adaptive} and healthcare~\cite{golmaei2021deepnote}. For example, in healthcare, GNNs can utilize information from patients and others with similar conditions to offer medical recommendations~\cite{min2024graph}. However, because users in these fields require reliable and accurate GNN predictions, the lack of transparency in the decision logic of GNNs has raised significant concerns about the credibility of GNN predictions. To address these concerns, the field of explainability of GNNs (XGNN) has emerged to enhance the transparency of the decision logic of GNNs and build up users' trust in GNN predictions. 
Existing XGNN methods~\cite{gnnexplainer_gegnn_2019,xgnn_tmegnn_2020,pgme_pgmegnn_2020} typically perturb the input graph structures to influence GNN predictions and extract the key subgraphs (a.k.a., \textit{explanation subgraphs}) that are most critical to support these predictions. These subgraphs are expected to represent the decision logic of GNNs to a certain extent, thereby effectively enhancing the predictions' reliability.

However, existing XGNN methods face two major limitations in real-world scenarios. 
\textbf{Limitation 1 (Incomplete Decision Logic):} Existing methods fail to capture the complete decision logic of GNNs across diverse distributions in the entire dataset's sample space, which overall may consist of all possible graph structures. These methods assume that the testing dataset (testing samples in the entire dataset) follows the same distribution as the training dataset (a.k.a., IID scenario) and focus solely on extracting the decision logic of GNNs in the training dataset. 
However, in real-world applications, out-of-distribution (OOD) scenarios are more prevalent~\cite{koch2024distribution,koch2022hidden}, where the testing dataset's distribution differs from that of the training dataset. In such cases, existing XGNN methods fail to provide reliable explanations because they focus either on OOD explanations that have different distributions from the training dataset or on the IID scenario. 
This limitation raises the demand of a novel \textit{comprehensive GNN explainer}, which needs to capture the complete decision logic of the target GNN across diverse distributions in the entire dataset's sample space and thus can: \textbf{(1)} generate reliable explanations in OOD scenarios, \textbf{(2)} help identify flaws in GNN decision logic when prediction errors occur, and \textbf{(3)} support GNN design improvements to mitigate errors in OOD scenarios.
\textbf{Limitation 2 (Strict Prerequisites):} Existing XGNN methods rely on strict prerequisites to achieve good performance, which can be divided into two aspects: \textbf{(1)} Most methods require GNN internal accessibility to extract the decision logic of GNNs. However, privacy protection laws and regulations often restrict such access, making these methods impractical for privacy-sensitive applications~\cite{miller2020adversarial}; \textbf{(2)} Several recent methods~\cite{wanggnninterpreter,chengenerating} generate learnable edge weights and require GNNs to use these weights for weighted message propagation. However, in critical fields like finance and healthcare, edge features (e.g., stock investment shares in finance and co-occurrence frequency of medical services in healthcare) are used to enhance data representation~\cite{li2022graph,xiong2021heterogeneous,zhu2023domain}. These edge features differ from learnable edge weights in both semantics and data formats, yet they share the same position in the input graph. Thus, these XGNN methods impose a prerequisite on dataset properties, requiring that datasets do not contain edge features. 

To address the abovementioned two major limitations in existing XGNN methods, we propose \Model, a novel c\textbf{O}mprehensive and \textbf{P}rerequisite-free \textbf{E}xplainer for graph \textbf{N}eural networks. 
Specifically, we propose the Non-Parametric Analysis Framework (NPAF), which infers the entire dataset's sample space from the training dataset samples and partitions this space into multiple environments. In addition, we propose the Graph Variational Generator (GVAG), which determines the sampling probabilities of graph structures by generating a large number of subgraphs in each environment during the training stage and analyzing their predictions. This enables GVAG to uncover the decision logic of the GNNs across a wide range of distributions in the sample space. To further enhance this uncovering process, GVAG incorporates node embeddings from other environments to actively construct OOD data. Compared to existing methods, \Model~can capture nearly complete decision logic of the GNNs, effectively addressing \textbf{Limitation 1}. Moreover, GVAG learns the correlations between embeddings and structure sampling probabilities, and directly samples explanation subgraphs from the sample space without accessing GNN internals or using edge weights, thereby eliminating the prerequisites required by existing methods and overcoming \textbf{Limitation 2}.

We summarize our main contributions as follows:
\textbf{(1)} We identify two major limitations in XGNN: (a) its inability to capture the complete decision logic, and (b) the strict prerequisites imposed on target GNNs and datasets. These limitations significantly impact XGNN research, highlighting the need for effective solutions;
\textbf{(2)} We propose \Model, a framework that infers and partitions the sample space of the entire dataset, enabling the exploration of the decision logic of GNNs across diverse distributions. In addition, \Model~learns the decision logic by sampling a large number of subgraphs and analyzing their predictions. This approach not only captures a more comprehensive decision logic than existing methods, but also removes the required strict prerequisites; and 
\textbf{(3)} Comprehensive experimental results demonstrate that \Model~not only effectively extracts explanation subgraphs in prerequisite-free scenarios where most existing methods are inapplicable, but also outperforms state-of-the-art (SOTA) methods in prerequisite-satisfied scenarios. \Model~efficiently generates reliable and accurate explanations across various distributions, showcasing its adaptability to real-world applications.

\vspace{-5pt}
\section{Related Work}

\textbf{XGNN.} 
The XGNN methods are initially inspired by computer vision techniques like Grad-CAM~\cite{explainability_mgcnn_2019}, and some methods also benefit from the explanatory power of GNN attention mechanisms~\cite{gat_2018}. The introduction of GNNExplainer~\cite{gnnexplainer_gegnn_2019} marked a shift towards perturbing graph structures to weight message propagation along edges and extracting explanation subgraphs based on changes in GNN outputs, as demonstrated by later advancements~\cite{qiu2024generating,chengenerating,cheninterpretable}. However, most existing methods rely on strict prerequisites to enable graph structure perturbation, which often restricts their practicality in real-world scenarios. Techniques like PGExplainer~\cite{parameterized_egnn_2020} enhance the understanding of how node embeddings correlate with node presence in explanation subgraphs, facilitating explanations post-learning without the need for fitting new instances~\cite{mixupexplainer_gegnnda_2023}. Nevertheless, these methods focus on learning the decision logic in the training dataset and thus fail to mine the complete decision logic of target GNNs, which causes them to provide unrelated explanations when the distribution of the input graph is different from that of the training dataset. 

\textbf{OOD Scenarios in XGNN.} 
In the XGNN field, critiques~\cite{d4explainer_idgnneddd_2023,chengenerating,fang2024evaluating,kubo2024xgexplainer,fang2024regularization} note that traditional methods generate OOD explanations, arguing that these explanations fail to accurately reflect the decision logic of GNNs. To address this issue, they propose generating explanations that align with the training dataset's distribution. However, this further aggravates the performance degradation of these methods in OOD scenarios. 
Some argue that an XGNN method only needs to be faithful to the decision logic of a well-trained GNN in the training dataset's distribution, and does not need to uncover the reasoning behind incorrect predictions in OOD scenarios. However, such opinion not only significantly limits the practical use of XGNN in critical fields where OOD scenarios are prevalent, but also prevents XGNN from contributing to improvements in GNN design.

\vspace{-5pt}
\section{Preliminaries}

\begin{figure}[t]
    \centering
    \includegraphics[width=\columnwidth]{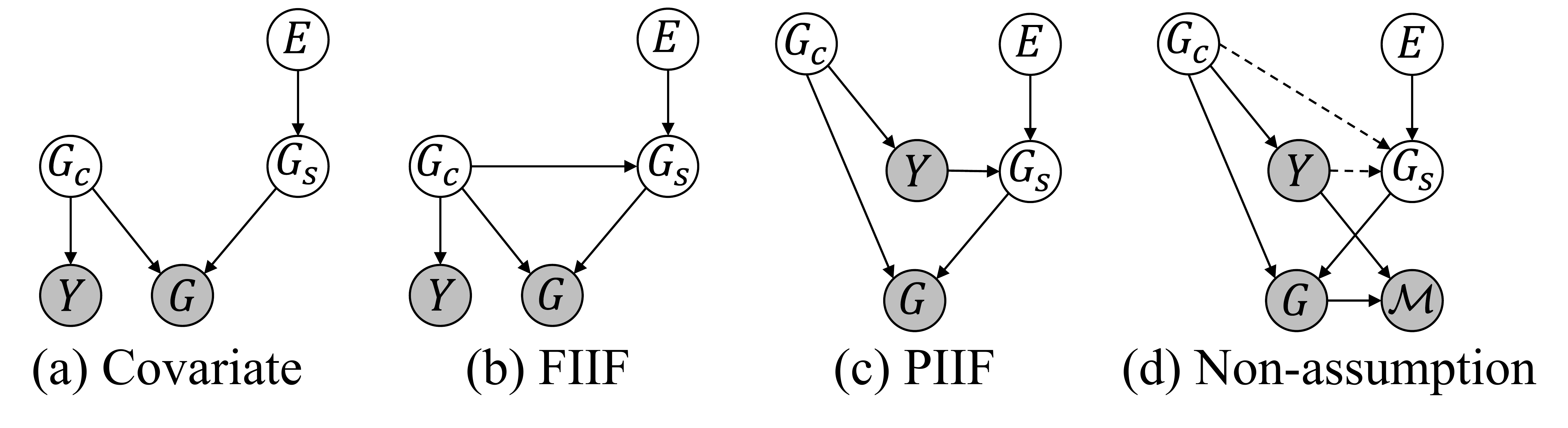}
    \vspace{-15pt}
    \caption{SCMs, where grey and white nodes indicate observable and unobservable variables, respectively. \(G_c\) represents a subgraph with a specific meaning and is used to determine label variables $Y$ of the input graph $G$. $G_s$ represents the part of $G$ influenced by environmental variables $E$. \(\mathcal{M}\) denotes the target GNN.}
    \label{fig:csm_example}
    \vspace{-5pt}
\end{figure}

\begin{figure*}[ht]
    \centering
    \includegraphics[width=0.86\textwidth]{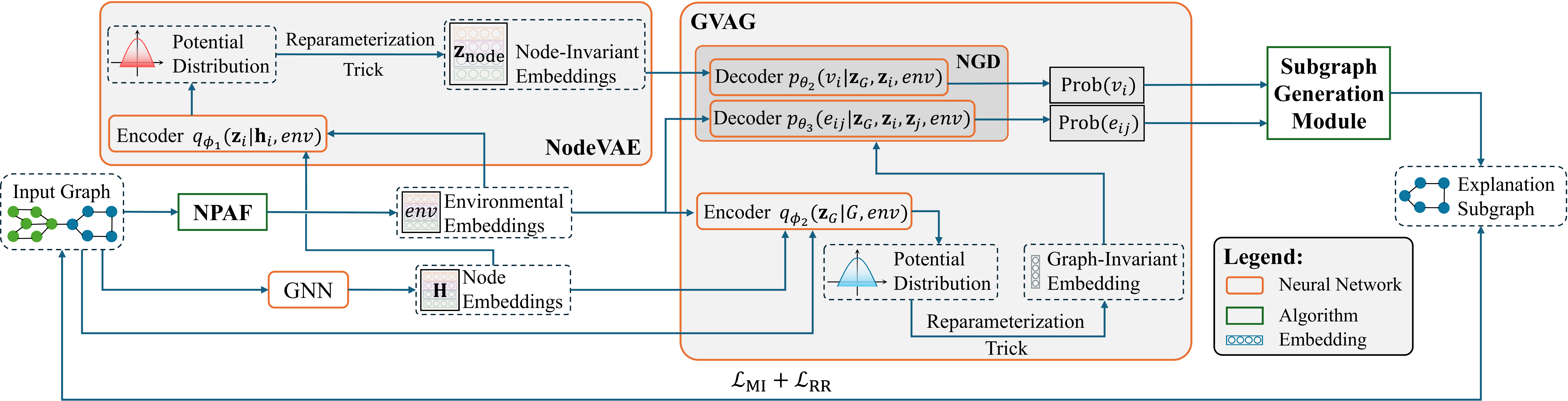}
    \vspace{-3pt}
    \caption{The overview of the proposed \Model~framework.}
    \label{fig:framework}
    \vspace{-6pt}
\end{figure*}

\textbf{Structural Causal Models (SCMs).}\label{sec:scm}
Following prior works~\cite{invariance_pmiboodg_2021,learning_ciroodgg_2022,ding2025few}, SCMs are used to delineate causal relationships among variables. 
Fig.~\ref{fig:csm_example} demonstrates three typical distribution shift assumptions in SCMs: Covariate, Fully Informative Invariant Features (FIIF), and Partially Informative Invariant Features (PIIF). Specifically, we propose a \textit{Non-assumption SCM} (shown in Fig.~\ref{fig:csm_example} (d)), which primarily explores potential causal relationships among variables \(G_c\), \(Y\), and \(G_s\). We can find that relying solely on \(Y\) and \(G\) is insufficient for precise causal analysis among these variables. Therefore, we perform direct statistical analyses to deduce environmental variables \(E\) and identify variables with distinct causal relationships to either \(Y\) or \(E\), effectively isolating relevant subsets of \(G_c\) and \(G_s\) and reducing the impact of spurious correlations. In addition, the target GNN \(\mathcal{M}\) has causal relationships with both \(G\) and \(Y\), indicating that when an XGNN method explains \(\mathcal{M}\)'s output, \(\mathcal{M}\) opens a backdoor path between the XGNN method and \(Y\). \textit{Therefore, XGNN methods also learn the relationships between \(Y\) and \(G_c\), and are susceptible to OOD scenarios.}

\textbf{Problem Definition.}
We focus on providing post-hoc instance-level explanations. Given a graph \( G = (\mathcal{V}, \mathcal{E}, \mathcal{X}) \) where \( \mathcal{V} \) denotes the vertices, \( \mathcal{E} \) denotes the edges, and \( \mathcal{X} \) is the node features, the target problem is defined as follows:
\begin{definition}[Mining complete decision logic of GNNs]
    Given a GNN model \( \mathcal{M} \) trained on the training dataset \( D_{\text{train}} \), consider an unseen testing dataset \( D_{\text{test}} \) where graphs exhibit distribution shifts from those in \( D_{\text{train}} \). For each \(G \in D_{\text{test}}\), the objective is to identify an explanation subgraph \( G_c \subseteq G\) that effectively elucidates the predictions made by \( \mathcal{M} \).
\end{definition}

Meanwhile, to eliminate the required prerequisites, we will not use learnable edge weights or access the GNN internal.

\section{Methodology of \Model}
In this section, we present a novel framework, comprehensive and prerequisite-free explainer for GNNs (\Model). 
Fig.~\ref{fig:framework} depicts the overview of \Model. A summary of all symbols used in this paper is provided in Appendix~\ref{app:symbol}. \Model~uses the Non-Parametric Analysis Framework (NPAF) to analyze potential environments in the entire dataset's sample space, based on training dataset samples. NPAF employs statistical methods following the non-assumption SCM and assigns environmental labels to the graphs in the training dataset. Next, NodeVAE and the Graph Variational Generator (GVAG) generate node and graph invariant embeddings in each environment, respectively. GVAG then samples explanation subgraphs from the sample space based on these invariant embeddings. \Model~adjusts the sampling probability of explanations by comparing the predictions, thus uncovering the decision logic of GNNs.  \footnote{Appendix can be found at https://github.com/zh2209645/OPEN}

\subsection{NPAF for Environmental Label Inference}\label{sec:npaf}
Taking the graph classification task as an example, NPAF determines the potential environmental label for each graph \( G_i \in D_\text{train} \) through the following procedure. 

\textbf{Obtain Structure-Based Embedding.}
NPAF leverages structure-based embeddings to infer environmental labels from the structural aspects of the training dataset. To generate these embeddings, we gather relevant structural information from the training dataset, such as node degrees and node categories. Using these details, we construct structure-based node features \( X_{str} \) and apply a Weisfeiler Leman (WL) kernel-based GNNs~\cite{wasserstein_wlgk_2019} to derive the structure-based embeddings for nodes \( H_{str} \). We then employ pooling layers to extract the structure-based embedding \( h_{G,i} \in H_{G} \) for \(G_i\).

\textbf{Infer Potential Environmental Label Based on Structure.}
We infer potential environmental labels for graphs based on a commonly adopted assumption in this field~\cite{discovering_irgnn_2022,learning_ciroodgg_2022}, which states that a graph can be divided into two independent components: \( G_c \), associated with the label \( Y \), and \( G_s \), influenced by the environment variable \( E \). Graphs affected by the same \( E \) are expected to share similar connection patterns during generation. 
Thus, potential environmental labels \( E \) can be inferred by analyzing and classifying graph structures. 
To assign structure-based environmental labels, we apply the K-Means algorithm to cluster structure-based embeddings \( H_G \) for \( G_i \in D_\text{train} \). The number of clusters, \( K \), determines the granularity of potential environments. A larger \( K \) indicates greater diversity and the presence of multiple potential environments, while a smaller \( K \) reflects less structural diversity. Based on the clustering results, each graph \( G_i \) is assigned an environmental label \( E^s_k \in E_{str} = \{ E^s_1, E^s_2, ..., E^s_K \} \).

\textbf{Identify Causal Structure in Graphs.} 
To identify nodes and edges in \( G_i \) that have a causal relationship with \( E^s_k \), we leverage the idea that graphs influenced by the same \( E \) exhibit similar structural patterns. Specifically, structure-based embeddings of nodes belonging to the \( G_s \) should show higher similarity within the same environment. This is because the embeddings \( H_{str} \) generated by the WL kernel effectively capture graph structures and reflect connection patterns. Since embeddings \( H_{str} \) are highly correlated with \( E \), they can be used to infer causal relationships between nodes and the environment. To pinpoint nodes irrelevant to the environment, we analyze the variance in their structure-based embeddings. For nodes sharing the same \( E^s_k \) and classification label \( Y \) within \( D_\text{train} \), we calculate the variance \( S^s_{k,Y} \) of their embeddings \( H_{str} \). By computing the gradient of \( S^s_{k,Y} \) with respect to each node's embedding \( h_i \in H_{str} \), we identify nodes with high gradients as irrelevant to the environment and group them into \( \mathcal{V}_c \). Nodes with lower gradients are assigned to \( G_s \), indicating their potential causal relationships with the environment \( E \). 

To determine edge significance in \( G_s \), we first generate a subgraph \( G'_i \) from the original graph \( G_i \) by dropping edges randomly. We then measure the distance change between the subgraph's embedding \( h'_{G,i} \) and the environment cluster centre of \( G_i \). A substantial change suggests an edge's importance to \( G_s \). By repeatedly testing each edge in \( G_i \), we identify and quantify the significance of edges. Critical edges are included in \( G_s \), while others are considered part of \( G_c \).

\textbf{Identify Causal Dimension in Node Features.}
To identify causal dimensions in node features, we observe that the distribution of feature dimensions affected by the same \( E \) should exhibit similarity across different node types and graph labels. This is because node features in distinct dimensions typically represent independent characteristics. Based on the research of existing SCMs~\cite{invariance_pmiboodg_2021}, nodes influenced by the same environment tend to show similar values in dimensions relevant to that environment. To determine these dimensions, we first group node features \( \mathcal{X} \) by their node type (if applicable) and the classification label \( Y \) of their corresponding graph. For each feature dimension, we compute its probability density across these groups. Using these densities, we construct a Jensen-Shannon (JS) divergence confusion matrix to evaluate the similarity of distributions. If a feature dimension demonstrates high similarity across different node types and classification labels, it suggests that this dimension is not specific to node type or graph label but is instead primarily influenced by environmental variables. Such dimensions are classified as \( Dim_{env} \), representing causal relationships to the environment. After identifying \( Dim_{env} \), we use a pooling layer to aggregate these dimensions from \( \mathcal{X} \) and generate graph-level features. These are then clustered using the K-Means to assign environmental labels \( E^f_k \in E_{feat} = \{E^f_1, E^f_2, ..., E^f_K\}\). However, the above procedure cannot completely exclude dimensions that are spuriously correlated with $Y$. To improve isolation, we assume dimensions influenced by the same \(E\) show consistent distributions across node types. We apply this method to nodes with the same \( E^f_k \), enhancing the precision of our analysis in identifying dimensions solely linked to node types.

Once environmental labels for graph structure (\( E^s_k \)) and node features (\( E^f_k \)) are inferred, we can use node features with varying environmental labels to create contrastive learning samples that diverge from the training dataset's distribution. This method refines the proposed \Model's ability to handle diverse distributions. In the inference stage, the learned cluster centres are used to predict environmental labels for new data.

\subsection{Invariant Learning and Subgraph Generation}
To explore how different environments affect graph structures and node features, we randomly initialize embeddings for environmental labels \(E_{str}\) and \(E_{feat}\), refining them through the training process. The target GNN \( \mathcal{M} \) is used to encode the graph \( G \) into node embeddings \( \mathbf{H} \), which are then aggregated via a pooling layer to form the graph embedding \( \mathbf{h}_G \).

\textbf{NodeVAE.} 
NodeVAE, built on conditional variational autoencoders, ensures that node features from different distributions are mapped to a unified embedding space. Firstly, it predicts label \( E^f_k \) using the NPAF module, and obtains the corresponding environmental embedding \( \mathbf{e}_i \) for node \( i \). Then, the NodeVAE encoder $g_{\phi_{1}}(\cdot)$, featuring a two-layers Multi-Layer Perceptrons (MLPs), processes node embedding \( \mathbf{h}_i \in \mathbf{H} \) and \( \mathbf{e}_i \) to determine the distribution \( q_{\phi_{1}}(\mathbf{z}_i|\mathbf{h}_i, env) \) as follows:
\begin{equation}
\boldsymbol{\mu}_i, \log (\boldsymbol{\sigma}_i^2) = g_{\phi_{1}}(\mathbf{h}_i, \mathbf{e}_i),
\end{equation}
where \(\phi_1\) represents the parameters of encoder models, \( \boldsymbol{\mu}_i \) and \( \log (\boldsymbol{\sigma}_i^2) \) denote the mean and log-variance of the node-invariant representation distribution of node $i$, respectively. \(env\) denotes the environmental embedding(s). The node potential invariant embedding $\mathbf{z}_i$ is sampled using reparameterization trick~\cite{kingma2013auto}:
% \vspace{-6.5pt}
\begin{equation}\label{equation:reparame}
\mathbf{z}_i = \boldsymbol{\mu}_i + \exp(1/2 \cdot \log (\boldsymbol{\sigma}_i^2)) \odot \boldsymbol{\epsilon}, \quad \boldsymbol{\epsilon} \sim \mathcal{N}(0, \mathbf{I}).
\end{equation}

NodeVAE's decoder \(f_{\theta_1}(\cdot)\) uses the learned distributions \(p_{\theta_1}(\mathbf{h}_i|\mathbf{z}_i, env)\) to reconstruct node embedding \(\hat{\mathbf{h}}_i\) from \(\mathbf{z}_i\) and \(\mathbf{e}_i\). The decoder, parameterized by \(\theta_1\), is followed by calculating the mean squared error between \(\mathbf{h}_i\) and \(\hat{\mathbf{h}}_i\), forming the reconstruction loss \(\mathcal{L}_\text{mse}\). Thus, NodeVAE's final loss is:
\begin{equation}
  \mathcal{L}_\text{NodeVAE} = \omega_\text{mse} \mathcal{L}_\text{mse} + \omega_\text{KL} D_{\text{KL}}\left(\mathcal{N}(\boldsymbol{\mu}_i, \boldsymbol{\sigma}_i^2) \| \mathcal{N}(0, \mathbf{I})\right),
\end{equation}
where \( \omega_\text{mse} \) and \( \omega_\text{KL} \) are used to balance the two terms, and \(D_{KL}(\cdot \parallel \cdot)\) donates the Kullback-Leibler (KL) divergence.

\textbf{GVAG.} GVAG can take into account the characteristics of different environments when sampling explanation subgraphs. GVAG first computes the graph's environmental embedding \(\mathbf{e}_G\) by averaging both \( E^s_k \) and \( E^f_k \) environmental embeddings. Then, GVAG's encoder $g_{\phi_2}(\cdot)$, consisting of a two-layers MLPs, leverages \(\mathbf{e}_G\) and the graph embedding \( \mathbf{h}_G \) to generate the graph-invariant embedding \(\mathbf{z}_G\) as follows:
\begin{equation}
\boldsymbol{\mu}_G, \log (\boldsymbol{\sigma}^2_G) = g_{\phi_2}(\mathbf{h}_G, \mathbf{e}_G),
\end{equation}
where \( \boldsymbol{\mu}_G \) and \( \log (\boldsymbol{\sigma}^2_G) \) denote the mean and log-variance of the graph-invariant representation distribution, respectively.
Then, the graph-invariant embedding $\mathbf{z}_G$ can be sampled with the reparameterization trick like Eq.~\ref{equation:reparame}.

The Novel Graph Decoder (NGD) in GVAG, unlike the GraphVAE~\cite{simonovsky2018graphvae}, which requires constructing and aligning a complete graph, directly models node and edge existence probabilities. Thus, GVAG improves scalability for large graphs and avoids the complexities of learnable edge weights, broadening its application range. The NGD consists of two decoders: \(f_{\theta_2}(\cdot)\) models the probability distributions of nodes (\(p_{\theta_2}(v_i | \mathbf{z}_G, \mathbf{z}_i, env)\)), and \(f_{\theta_3}(\cdot)\) handles the distributions for edges (\(p_{\theta_3}(e_{ij}|\mathbf{z}_G, \mathbf{z}_i, \mathbf{z}_j, env)\)), both using \(\mathbf{z}_G\), node-invariant embeddings \(\mathbf{z}_\text{node}\), and \(\mathbf{e}_G\). The specific formulas are as follows:
\begin{equation}
{\textstyle \text{Prob}({v_i}) = f_{\theta_2}(\mathbf{z}_G, \mathbf{z}_i, \mathbf{e}_G), \text{Prob}({e_{ij}}) = f_{\theta_3}(\mathbf{z}_G, \mathbf{z}_i, \mathbf{z}_j, \mathbf{e}_G),}
\end{equation}
where $\text{Prob}({v_i})$ and $\text{Prob}({e_{ij}})$ denote the probabilities that node $i$ and edge $e_{ij}$ exist on the explanation subgraph, respectively. By leveraging node-invariant and graph-invariant embeddings instead of standard node and graph embeddings, our subgraph generation method efficiently handles varying data distributions. In addition, GVAG employs another NGD instance to establish causal relationships between nodes, edges, and labels \(Y\), serving as a regularization term to improve the refinement of environmental embeddings.

GVAG utilizes direct reconstruction loss to refine graph generation by optimizing node and edge existence probabilities. This method maximizes the mutual information between the explanation subgraph and the predicted label, given the input graph, as follows:
\begin{align}
\mathcal{L}_\text{MI} = &{\textstyle - \sum\nolimits_{G_c} \mathbf{1}[Y = \widetilde{Y}] \log (\text{Prob}(G_c))} \nonumber\\
&{\textstyle + \sum\nolimits_{G_c} \mathbf{1}[Y\ne \widetilde{Y}] \log (\text{Prob}(G_c)),}
\end{align}
where $G_c$ denotes the generated subgraph, $\mathbf{1}[\cdot]$ is an indicator function, 
and $Y$ and $\widetilde{Y}$ are the outputs of the GNN for the original graph $G$ and the subgraph $G_c$, respectively. 
Furthermore, to optimize node and edge existence probabilities in explanations, we leverage the Reconstruction Regularization Loss (\(\mathcal{L}_{\text{RR}}\)). This approach adjusts probabilities by comparing the GNN prediction loss between the original graph \(G_i \in D_{\text{train}} \) and its explanation \(G_{c,i}\), effectively emphasizing crucial substructures in the subgraph. The specific formula is as follows:
\begin{equation}
{\textstyle \mathcal{L}_{\text{RR}} = \sum\nolimits_i (\mathcal{L}_\text{diff}^i \cdot \text{Prob}({G_{c,i}})),~
\mathcal{L}_\text{diff}^i = \mathcal{L}({G_{c,i}}) - \mathcal{L}({G_i}),}
\end{equation}
where $\mathcal{L}(\cdot)$ is the loss function used to train $\mathcal{M}$.

\textbf{Subgraph Generation.}
By utilizing the mean-field variational approximation theory~\cite{mean_ftgnngp_2018,gnnexplainer_gegnn_2019}, we complete the subgraph generation process based on the probabilities of nodes and edges. 
\Model~dynamically constructs the explanation by adding nodes and edges based on \( 
\text{Prob}({v_i}) \) and \( \text{Prob}({e_{ij}}) \). \textit{Pseudocode and computational complexity analysis can be found in the Appendix~\ref{app:recon_algor}.} After subgraph \(G_c\) is generated, the probability of \( G_c \) is determined using the following equation:
\begin{equation}
    {\textstyle \text{Prob}(G_c) = {\textstyle \prod_{v_i \in G_c}}\text{Prob}({v_i}) \cdot {\textstyle \prod_{e_{ij} \in G_c}}\text{Prob}({e_{ij}}).}
\end{equation}
This probabilistic method lets users adjust the scale and density of subgraphs, significantly improving explanation quality. 

\subsection{Regularization Terms}
\textbf{Causal Structure Regularization.} We employ binary cross-entropy to optimize environmental embeddings and predict causal relationships between graph structures and labels \(Y\). The calculation formula of \(\mathcal{R}_\text{causal}\) is computed as follows:
\begin{equation}
\resizebox{\linewidth}{!}{$
\begin{aligned}
    \mathcal{R}_\text{causal} = - \mathbb{E}_{G} [ &\mathbb{E}_{v \in \mathcal{V}} [ y_v \cdot \log(\sigma(z_v)) + (1 - y_v) \cdot \log(1 - \sigma(z_v)) ] \\
    + & \mathbb{E}_{e \in \mathcal{E}} [ y_e \cdot \log(\sigma(z_e)) + (1 - y_e) \cdot \log(1 - \sigma(z_e)) ] ],
\end{aligned}
$}
\end{equation}
where \( \sigma(\cdot) \) denotes the sigmoid function, and \( z_v \) and \( z_e \) represent the prediction logits for the node \( v \) and edge \( e \), respectively. \( y_v = 1 \) if node \( v \) is part of \( G_c \), otherwise \( y_v = 0 \). Similarly, \( y_e \) follows the same pattern as \( y_v \).

\textbf{Hinge Regularization.} We introduce hinge regularization to constrain \Model~learning process, ensuring the empirical loss of \( G_c \) is smaller than \( G_s \). The formula is as follows:
\begin{equation}
    {\textstyle \mathcal{R}_\text{hinge} = \mathbb{E}_{\mathcal{L}(G_s) > \mathcal{L}(G_c)}\sum{\mathcal{L}(G_s)}. }
\end{equation}

\textbf{Subgraph Node Count Regularization.} We design constraints on the compactness of subgraphs. Specifically, given the loss function outputs for original graphs \( \mathcal{L}({G}) \) and explanation subgraphs \( \mathcal{L}({G_c}) \), subgraph node counts \( n_{\text{sub}} \), and prior node counts \( n_{\text{prior}} \), this loss for instance \( i \) is defined as:
\begin{equation}
\mathcal{R}_\text{subg\_node}^i = 
\begin{cases} 
\left(\frac{1}{\mathcal{L}_\text{diff}^i + \epsilon}\right) \left(\frac{n_{\text{sub},i} - n_{\text{prior},i}}{n_{\text{prior},i}}\right), & \text{if } \mathcal{L}_\text{diff}^i > 0 \\
\left(\frac{1}{\mathcal{L}_\text{diff}^i + \epsilon}\right) \left(\frac{1}{n_{\text{sub},i}}\right), & \text{if } \mathcal{L}_\text{diff}^i \leq 0 
\end{cases}
\end{equation}
where $\epsilon$ is a small constant to prevent division by zero, the overall regularization $\mathcal{R}_\text{subg\_node}$ is computed as the average of the \( \mathcal{R}_\text{subg\_node}^i \) for all graphs in \( D_\text{train} \).

\begin{table}[t]
\centering

\resizebox{0.69\linewidth}{!}{%
\begin{tabular}{lclcl}
\hline \hline
Dataset & \multicolumn{2}{c}{Cora} & \multicolumn{2}{c}{Motif} \\ \hline
Input(X) & \multicolumn{2}{c}{Scientific   publications} & \multicolumn{2}{c}{Motif-base graphs} \\
Prediction(Y) & \multicolumn{2}{c}{Publication classes} & \multicolumn{2}{c}{Motifs} \\
\#Subgraphs & \multicolumn{2}{c}{19,793} & \multicolumn{2}{c}{30,000} \\
\#Nodes & \multicolumn{2}{c}{8,896,055} & \multicolumn{2}{c}{785,320} \\
\#Edges & \multicolumn{2}{c}{64,479,758} & \multicolumn{2}{c}{2,085,430} \\
Domain & \multicolumn{2}{c}{Word/Degree} & \multicolumn{2}{c}{Basis/Size} \\
\#Domains & \multicolumn{2}{c}{218/102} & \multicolumn{2}{c}{5/5} \\
Shift Type & \multicolumn{1}{l}{Covariate} & Concept & \multicolumn{1}{l}{Covariate} & Concept \\
\#Environments &  \\ 
(train:val:test) & \multicolumn{1}{l}{10:1:1} & 3:1:1 & \multicolumn{1}{l}{3:1:1} & 3:1:1 \\ \hline \hline
\end{tabular}%
}
\caption{Dataset statistics. }
\label{tab:dataset_statis}
\end{table}

\begin{table}[]
\centering

\resizebox{\columnwidth}{!}{%
\begin{tabular}{lllllll|llllll}
\hline \hline
Dataset & \multicolumn{6}{c|}{Cora} & \multicolumn{6}{c}{Motif} \\ \hline
Shift Domain & \multicolumn{3}{c|}{Degree} & \multicolumn{3}{c|}{Word} & \multicolumn{3}{c|}{Basis} & \multicolumn{3}{c}{Size} \\
 & \( \text{fid}_{+} \) \(\uparrow\) & \( \text{fid}_{-} \) \(\downarrow\) & \multicolumn{1}{l|}{\( \text{GEF} \) \(\downarrow\)} & \( \text{fid}_{+} \) \(\uparrow\) & \( \text{fid}_{-} \) \(\downarrow\) & \( \text{GEF} \) \(\downarrow\) & \( \text{fid}_{+} \) \(\uparrow\) & \( \text{fid}_{-} \) \(\downarrow\) & \multicolumn{1}{l|}{\( \text{GEF} \) \(\downarrow\)} & \( \text{fid}_{+} \) \(\uparrow\) & \( \text{fid}_{-} \) \(\downarrow\) & \( \text{GEF} \) \(\downarrow\) \\ \hline
GMT-SAM & 0.0563 & 0.5676 & \multicolumn{1}{l|}{0.0777} & 0.0645 & 0.6980 & 0.0931 & 0.0063 & \textbf{0.0031} & \multicolumn{1}{l|}{\textbf{0.0037}} & \textbf{0.0301} & 0.2148 & 0.0034 \\
\Model & \textbf{0.5704} & \textbf{0.3318} & \multicolumn{1}{l|}{\textbf{0.0279}} & \textbf{0.6932} & \textbf{0.2674} & \textbf{0.0164} & \textbf{0.0110} & 0.1930 & \multicolumn{1}{l|}{0.0540} & 0.0120 & \textbf{0.1123} & \textbf{0.0018} \\ \hline \hline
\end{tabular}%
}
\caption{The comparison of \Model~and the baseline under prerequisite-free scenarios. }
\label{tab:prerequisite_free}
\end{table}

\subsection{Other Loss Functions}
\textbf{Contrastive Loss.} We implement a specialized contrastive learning loss in \Model~to bolster its resistance to environmental interference. Given a set of original graph embeddings and their perturbed counterparts grouped by class labels, the formula for the contrastive loss $\mathcal{L}_\text{CON}$ is given by:
\begin{equation}
    {\textstyle \mathcal{L}_\text{CON} = -\log\left(\frac{\sum \exp(s_{\text{intra}})}{\sum \exp(s_{\text{intra}}) + \sum \exp(s_{\text{inter}}) + \epsilon}\right),}
\end{equation}
where \( s_{\text{intra}} \) and \( s_{\text{inter}} \) denote the similarity scores among graph-invariant embeddings within the same class and across different classes, respectively. $\epsilon$ is used to avoid calculation issues.

\textbf{Last Action Rewards (LAR).} Inspired by reinforcement learning, we introduce reward functions for mining explanations in \Model. Rewards are assigned based on whether the prediction accuracy of explanations improves post-update; penalties apply if it worsens. The LAR is defined as:
\begin{equation}
    \text{LAR} = (\mathbb{E}(\mathcal{L}_\text{diff}) - \mathbb{E}(\mathcal{L}_\text{old\_diff})) \cdot \mathbb{E}_{G_c}{\text{Prob}(G_C)},
\end{equation}
where $\mathbb{E}(\mathcal{L}_\text{diff})$ and $\mathbb{E}(\mathcal{L}_\text{old\_diff})$ are the expectation of $\mathcal{L}_\text{diff}$ in this epoch and last epoch, respectively.

\textbf{Final Loss of \Model.} The final optimization function for our proposed method is as follows:
\begin{equation}
\small
\begin{split}
    \mathcal{L}_\text{final} = \omega_\text{NodeVAE}\mathcal{L}_\text{NodeVAE} + \omega_\text{RECON}(\mathcal{L}_\text{MI} + \mathcal{L}_\text{RR}) + \omega_\text{CON}\mathcal{L}_\text{CON} \\ +   \omega_\text{LAR}\text{LAR}+\mathcal{R}_\text{causal}+\mathcal{R}_\text{hinge}+\mathcal{R}_\text{subg\_node},
\end{split}
\end{equation}
where \(\omega_\text{NodeVAE}\), \(\omega_\text{RECON}\), \(\omega_\text{CON}\) and \(\omega_\text{LAR}\) are hyper-parameters.

\begin{table*}[t]
\centering

\resizebox{0.64\textwidth}{!}{%
\begin{tabular}{cccllllll|llllll}
\hline \hline
\multicolumn{1}{l}{} & \multicolumn{1}{l}{} & \multicolumn{1}{l}{} &  & \multicolumn{5}{c|}{L\&P XGNN (OOD sensitive)} & \multicolumn{6}{c}{Ideal Performance for Reference (OOD insensitive)} \\
\multicolumn{1}{l}{Shift Type} & \multicolumn{1}{l}{Dataset} & \multicolumn{2}{l}{Shift Domain} & PGExp & MixupExp & GMT-SAM & ProxyExp & \Model & GradCAM & ATT & GNNExp & PGMExp & CF-GNNExp & KRCW \\ \hline
\multirow{12}{*}{Covariate Shift} & \multirow{6}{*}{Cora} & \multirow{3}{*}{Degree} & \( \text{fid}_{+} \) \(\uparrow\) & 0.2554 & 0.2252 & 0.0563 & 0.2500 & \textbf{0.5704} (+123.3\%) & 0.0038 & 0.5753 & {\ul 0.6045} & 0.5871 & 0.3807 & 0.5078 \\
 &  &  & \( \text{fid}_{-} \) \(\downarrow\) & 0.5009 & 0.5281 & 0.5676 & 0.5559 & \textbf{0.3318} (+33.76\%) & 0.5851 & 0.3283 & 0.1398 & 0.2284 & 0.3599 & {\ul 0.0391} \\
 &  &  & \( \text{GEF} \) \(\downarrow\) & 0.0678 & 0.0716 & 0.0777 & 0.0691 & \textbf{0.0279} (+58.85\%) & 0.0805 & 0.0323 & 0.0139 & 0.0254 & 0.0432 & {\ul 0.0052} \\ \cline{3-15} 
 &  & \multirow{3}{*}{Word} & \( \text{fid}_{+} \) \(\uparrow\) & 0.2018 & 0.1669 & 0.0645 & 0.2020 & \textbf{0.6932} (+243.2\%) & 0.0005 & 0.1862 & {\ul 0.7435} & 0.6961 & 0.3306 & 0.5234 \\
 &  &  & \( \text{fid}_{-} \) \(\downarrow\) & 0.5882 & 0.6239 & 0.6980 & 0.6266 & \textbf{0.2674} (+54.54\%) & 0.6956 & 0.6292 & 0.1225 & 0.4147 & 0.3573 & {\ul 0.0859} \\
 &  &  & \( \text{GEF} \) \(\downarrow\) & 0.0710 & 0.0765 & 0.0931 & 0.0676 & \textbf{0.0164} (+75.74\%) & 0.0960 & 0.0844 & 0.0074 & 0.0498 & 0.0301 & {\ul 0.0055} \\ \cline{2-15} 
 & \multirow{6}{*}{Motif} & \multirow{3}{*}{Basis} & \( \text{fid}_{+} \) \(\uparrow\) & 0.0133 & \textbf{0.0163} & 0.0063 & N/A & 0.0110 & 0.2097 & 0.1390 & 0.1713 & {\ul 1.0000} & 0.1727 & {\ul 1.0000} \\
 &  &  & \( \text{fid}_{-} \) \(\downarrow\) & 0.1950 & 0.1956 & \textbf{0.0031} & N/A & 0.1930 & 0.0343 & 0.0707 & 0.1713 & 0.1950 & 0.1687 & {\ul 0.0008} \\
 &  &  & \( \text{GEF} \) \(\downarrow\) & 0.0559 & 0.0542 & \textbf{0.0037} & N/A & 0.0540 & 0.0389 & 0.0181 & 0.0498 & 0.0543 & 0.0498 & {\ul 0.0000} \\ \cline{3-15} 
 &  & \multirow{3}{*}{Size} & \( \text{fid}_{+} \) \(\uparrow\) & \textbf{0.0410} & 0.0387 & 0.0301 & N/A & 0.0120 & 0.1843 & 0.0490 & 0.0510 & {\ul 1.0000} & 0.0483 & 0.9941 \\
 &  &  & \( \text{fid}_{-} \) \(\downarrow\) & 0.2540 & 0.2483 & 0.2148 & N/A & \textbf{0.1123} (+47.72\%) & 0.2673 & 0.0777 & 0.0513 & 0.1240 & 0.0530 & {\ul 0.0004} \\
 &  &  & \( \text{GEF} \) \(\downarrow\) & 0.0057 & 0.0053 & 0.0034 & N/A & \textbf{0.0018} (+47.06\%) & 0.2054 & 0.0146 & 0.0001 & 0.0020 & 0.0001 & {\ul 0.0000} \\ \hline
\multirow{12}{*}{Concept Shift} & \multirow{6}{*}{Cora} & \multirow{3}{*}{Degree} & \( \text{fid}_{+} \) \(\uparrow\) & 0.1850 & 0.1848 & 0.0605 & 0.1660 & \textbf{0.6146} (+232.2\%) & 0.0306 & 0.1850 & {\ul 0.7464} & 0.6233 & 0.3276 & 0.4297 \\
 &  &  & \( \text{fid}_{-} \) \(\downarrow\) & 0.5376 & 0.5379 & 0.5820 & 0.5539 & \textbf{0.2843} (+47.12\%) & 0.6054 & 0.5664 & 0.1080 & 0.2944 & 0.3198 & {\ul 0.1016} \\
 &  &  & \( \text{GEF} \) \(\downarrow\) & 0.0741 & 0.0725 & 0.0805 & 0.0558 & \textbf{0.0196} (+64.87\%) & 0.0846 & 0.0781 & {\ul 0.0065} & 0.0323 & 0.0308 & 0.0176 \\ \cline{3-15} 
 &  & \multirow{3}{*}{Word} & \( \text{fid}_{+} \) \(\uparrow\) & 0.1513 & 0.1641 & 0.0512 & 0.1555 & \textbf{0.6166} (+275.7\%) & 0.0131 & 0.1335 & {\ul 0.7114} & 0.6242 & 0.3040 & 0.4766 \\
 &  &  & \( \text{fid}_{-} \) \(\downarrow\) & 0.5541 & 0.5385 & 0.5777 & 0.5523 & \textbf{0.2821} (+47.61\%) & 0.6133 & 0.5687 & {\ul 0.0942} & 0.3357 & 0.2975 & 0.1484 \\
 &  &  & \( \text{GEF} \) \(\downarrow\) & 0.0791 & 0.0754 & 0.0793 & 0.0554 & \textbf{0.0200} (+63.90\%) & 0.0873 & 0.0807 & {\ul 0.0052} & 0.0394 & 0.0263 & 0.0054 \\ \cline{2-15} 
 & \multirow{6}{*}{Motif} & \multirow{3}{*}{Basis} & \( \text{fid}_{+} \) \(\uparrow\) & \textbf{0.0323} & 0.0282 & 0.0293 & N/A & 0.0250 & 0.0925 & 0.0282 & 0.0255 & {\ul 1.0000} & 0.0223 & {\ul 1.0000} \\
 &  &  & \( \text{fid}_{-} \) \(\downarrow\) & \textbf{0.0473} & 0.0905 & 0.1305 & N/A & 0.0768 & 0.0188 & 0.2688 & 0.0227 & 0.1273 & 0.0238 & {\ul 0.0000} \\
 &  &  & \( \text{GEF} \) \(\downarrow\) & \textbf{0.0302} & 0.0380 & 0.0354 & N/A & 0.0458 & 0.0334 & 0.1367 & 0.0050 & 0.0661 & 0.0048 & {\ul 0.0000} \\ \cline{3-15} 
 &  & \multirow{3}{*}{Size} & \( \text{fid}_{+} \) \(\uparrow\) & 0.1698 & \textbf{0.1858} & 0.0660 & N/A & 0.0730 & 0.3805 & 0.1077 & 0.0912 & {\ul 1.0000} & 0.0798 & {\ul 1.0000} \\
 &  &  & \( \text{fid}_{-} \) \(\downarrow\) & 0.5392 & 0.4990 & 0.4559 & N/A & \textbf{0.3688} (+19.11\%) & 0.0370 & 0.4090 & 0.0893 & 0.3763 & 0.0925 & {\ul 0.0027} \\
 &  &  & \( \text{GEF} \) \(\downarrow\) & 0.1044 & 0.0948 & 0.0902 & N/A & \textbf{0.0894} (+0.887\%) & 0.0593 & 0.0983 & 0.0043 & 0.0707 & 0.0050 & {\ul 0.0001} \\ \hline \hline
\end{tabular}%
}
\caption{The comparison of \Model~and baselines under prerequisite-satisfied scenarios. \(\uparrow\) and \(\downarrow\) represent that higher is better and lower is better, respectively. \textbf{Bold} indicates the best results among all methods, and \uline{underline} indicates the ideal performance, which is not affected by OOD scenarios, for reference.}
\label{tab:ood_comparison}
\end{table*}

\section{Experiments}
In this section, we conduct a series of experiments to comprehensively evaluate the effectiveness of \Model. The primary objectives are to assess fidelity, robustness and efficiency across various OOD scenarios, answering the following research questions: \textbf{RQ1:} How does \Model~perform in terms of fidelity and robustness compared to baseline methods in prerequisite-free and prerequisite-satisfied scenarios? \textbf{RQ2:} How does \Model~perform in terms of explanation subgraph quality and interpretation efficiency compared to baseline methods? \textbf{RQ3:} How do different modules within \Model~contribute to its performance? More experiments can be found in Appendix~\ref{app:exp_sec}.

\textbf{Datasets.} 
We use the Graph Out-of-Distribution Benchmark (GOOD)~\cite{good_goodb_2022} to provide OOD scenarios, and select two widely used datasets, Cora and Motif, from it. Each dataset includes two shift types (covariate and concept) and two shift domains, resulting in a total of eight cases (2 datasets \(\times\) 2 types \(\times\) 2 domains). Table~\ref{tab:dataset_statis} provides detailed dataset statistics. Specifically, Cora is a complex real-world dataset with high node and edge densities and diverse distributions. In contrast, Motif is a simpler artificial dataset with a lower average node degree and fewer distributions. However, as an artificially generated dataset, Motif introduces a two-level OOD issue in the basis domain: the first-level OOD issue results from distribution shifts in the base part of the graphs, while the second-level OOD issue arises from distribution shifts in the motif part. This combination encompasses nearly all factors influencing XGNN performance, supporting a balanced and comprehensive evaluation of \Model.

\begin{table*}[]
\centering

\resizebox{0.62\linewidth}{!}{%
\begin{tabular}{ccllllll|llllll}
\hline \hline
\multicolumn{1}{l}{} & \multicolumn{1}{l}{} &  & \multicolumn{5}{c|}{L\&P XGNN (OOD sensitive)} & \multicolumn{6}{c}{Ideal Performance for Reference (OOD insensitive)} \\
\multicolumn{1}{l}{Dataset} & \multicolumn{2}{l}{Shift Domain} & PGExp & MixupExp & GMT-SAM & ProxyExp & \Model & GradCAM & ATT & GNNExp & PGMExp & CF-GNNExp & KRCW \\ \hline
\multirow{6}{*}{Cora} & \multirow{3}{*}{Degree} & \( \rho_v \) \(\downarrow\) & 0.5215 & 0.5250 & 0.2352 & 0.6130 & 0.4338 & 0.0122 & 0.3294 & 0.4681 & 0.3527 & 0.8967 & 0.9657 \\
 &  & \( \rho_e \) \(\downarrow\) & 0.2361 & 0.2092 & 0.0450 & 0.2255 & 0.2092 & 0.0070 & 0.1930 & 0.2340 & 0.2862 & 0.5069 & 0.9928 \\
 &  & \( T \) \(\downarrow\) & 48.065 & 8.1771 & 2.0600 & 5.7219 & 7.3380 & 0.2172 & 0.6060 & 24.518 & 88.125 & 18.918 & 9073.4 \\ \cline{2-14} 
 & \multirow{3}{*}{Word} & \( \rho_v \) \(\downarrow\) & 0.4910 & 0.4664 & 0.2854 & 0.5651 & 0.5412 & 0.0030 & 0.1540 & 0.3973 & 0.1765 & 0.9318 & 0.8714 \\
 &  & \( \rho_e \) \(\downarrow\) & 0.1974 & 0.1679 & 0.0415 & 0.1727 & 0.3613 & 0.0013 & 0.0638 & 0.1812 & 0.1261 & 0.5050 & 0.9070 \\
 &  & \( T \) \(\downarrow\) & 48.9642 & 33.4767 & 2.9617 & 13.5386 & 27.9979 & 0.2177 & 0.6217 & 25.0788 & 112.5488 & 18.6097 & 79821.2043 \\ \hline
\multirow{6}{*}{Motif} & \multirow{3}{*}{Basis} & \( \rho_v \) \(\downarrow\) & 0.2771 & 0.2695 & 0.5259 & N/A & 0.3445 & 0.4999 & 0.5432 & 0.9276 & 0.0689 & 0.9229 & 0.9916 \\
 &  & \( \rho_e \) \(\downarrow\) & 0.1659 & 0.1659 & 0.1890 & N/A & 0.1637 & 0.3848 & 0.2346 & 0.5047 & 0.0000 & 0.4981 & 1.0000 \\
 &  & \( T \) \(\downarrow\) & 0.9975 & 1.2111 & 1.8180 & N/A & 8.6190 & 6.0286 & 0.7405 & 19.3248 & 74.9083 & 17.0641 & 37.9837 \\ \cline{2-14} 
 & \multirow{3}{*}{Size} & \( \rho_v \) \(\downarrow\) & 0.2561 & 0.2544 & 0.4297 & N/A & 0.2253 & 0.3198 & 0.5246 & 0.9020 & 0.0247 & 0.9081 & 0.9962 \\
 &  & \( \rho_e \) \(\downarrow\) & 0.1524 & 0.1524 & 0.1726 & N/A & 0.0981 & 0.2308 & 0.2015 & 0.4802 & 0.0007 & 0.4933 & 0.9998 \\
 &  & \( T \) \(\downarrow\) & 1.0011 & 4.5940 & 1.8287 & N/A & 3.6504 & 6.1518 & 0.7527 & 28.5231 & 90.2539 & 17.4845 & 198.1647 \\ \hline \hline
\end{tabular}%
}
\caption{The statistics of comparison on covariate shift scenarios.}
\label{tab:covariate_info}
\end{table*}

\textbf{Baselines.} 
Considering dataset compatibility and GNN requirements, we select several Learning \& Prediction (L\&P) type XGNN methods as SOTA baselines, including: \textbf{PGExp}~\cite{parameterized_egnn_2020}, \textbf{MixupExp}~\cite{mixupexplainer_gegnnda_2023}, \textbf{GMT-SAM}~\cite{cheninterpretable}, and \textbf{ProxyExp}~\cite{chengenerating}. This type of XGNN method is notably efficient and capable of providing explanations for new instances immediately after training. In addition, we include XGNN methods that do not have a learning phase and must fit each instance individually to identify the explanation subgraphs. Although these methods cannot learn the decision logic of the GNNs and are inefficient, they are unaffected by OOD scenarios and can, therefore, serve as an \textit{ideal performance for reference}, representing the fidelity of the generated explanation subgraphs when complete decision logic is fully learnt. These include \textbf{GradCAM}~\cite{explainability_mgcnn_2019}, \textbf{ATT}~\cite{gat_2018}, \textbf{GNNExp}~\cite{gnnexplainer_gegnn_2019}, \textbf{PGMExp}~\cite{pgme_pgmegnn_2020}, \textbf{CF-GNNExp}~\cite{cf_cegnn_2022}, and \textbf{KRCW}~\cite{qiu2024generating}. We implement these methods using well-established libraries like torch-geometric~\cite{FeyLenssen2019} and DIG~\cite{JMLRv22210343}, among other reliable sources.

\textbf{Setup.} To evaluate the performance of ATT, we use a 3-layer GAT network~\cite{gat_2018} as backbone GNN $\mathcal{M}$. Other experimental setup, including the hardware and software platform, as well as the hyper-parameter settings, can be found in the \textbf{Appendix~\ref{app:exp_setup}}. Experiments on the hyper-parameter sensitivity can be found in the \textbf{Appendix~\ref{app:hyperparam_sensiti}}.

\textbf{Evaluation Metrics.} We select five widely used metrics to evaluate XGNN methods comprehensively~\cite{graphframex_tseemgnn_2022,agarwal2023evaluating}: 
\textbf{(1) Negative Fidelity (\(\text{fid}_{-}\))} measures the inconsistency between the predicted labels of \(G_c\) and \(G\) (lower is better). This metric highlights the relevance of the explanation subgraph to the model's decision logic and is \textit{the most important metric} in relative terms.
% The calculation is defined as:
% \centerline{
% \(
%     \text{fid}_{-} = 1 - \mathbb{E}_{G \in D_{test}} \mathbf{1}(Y_{G_c} = Y),
% \)
% }
% where \( Y_{G_c} \) and \( Y \) represent the classification labels of \( G_c \) and \( G \), respectively.
\textbf{(2) Positive Fidelity (\(\text{fid}_{+}\))} evaluates the inconsistency between predicted labels of \(G_s\) and $G$ (higher is better). 
% The calculation is defined by the following formula:
% \centerline{
% \(
%     \text{fid}_{+} = 1 - \mathbb{E}_{G \in D_{test}} \mathbf{1}(Y_{G_s} = Y),
% \)
% }
% where \( Y_{G_s} \) represents the classification labels of \( G_s \).
\textbf{(3) Unfaithfulness (\(\text{GEF}\))} is quantified using KL divergence between prediction distributions of \(G_c\) and \(G\), which provides insights into the reliability of explanations. 
% The calculation is defined by the following formula:
% \centerline{
% \(
%     \text{GEF}(Y, Y_{G_c}) = 1 - \exp(-D_{\text{KL}}(Y \parallel Y_{G_c})).
% \)
% }
% where \( D_{\text{KL}}(\cdot \parallel \cdot) \) represents the KL divergence.
Besides, \textbf{(4) Node Density (\( \rho_v \))} and \textbf{(5) Edge Density (\( \rho_e \))} are given to determine the compactness of explanations. Method complexity is evaluated by measuring the \textbf{time (\( T \))} required to generate explanations for $100$ samples, expressed in seconds. 

\subsection{Performance Comparison (RQ1)}
We evaluate the fidelity and robustness of \Model~and baselines under two scenarios: (1) prerequisite-free, where access to GNN internals and the use of learnable edge weights are prohibited; and (2) prerequisite-satisfied, where these operations are permitted. In the prerequisite-satisfied scenario, XGNN methods can perturb the internal dataflows of GNNs, and generate learnable edge weights and use them as part of the input when the dataset does not contain edge features.

\textbf{Prerequisite-Free Comparisons.}  
Because PGExp and MixupExp require perturbing the internal dataflows of GNNs, and ProxyExp relies on learnable edge weights as part of the input, only GMT-SAM is applicable in prerequisite-free scenarios. Thus, GMT-SAM serves as the sole baseline for \Model~in covariate shift settings. As shown in Table~\ref{tab:prerequisite_free}, \Model~outperforms GMT-SAM consistently. On the Cora dataset, \Model~shows a 356.26\% average improvement across all metrics in both degree and word domains. On the Motif dataset, \Model~performs comparably to GMT-SAM, primarily due to its lower-than-expected performance in the basis domain. This is because the two-level OOD issue does not align with the SCM used by \Model, which limits NPAF's ability to partition the sample space, resulting in its performance degradation. Overall, these results indicate that by inferring and partitioning the entire dataset's sampling space, \Model~captures a more complete GNN decision logic in complex datasets like Cora, leading to greater performance improvements than in simpler datasets like Motif. This makes \Model~more effective for complex datasets. In addition, by adopting a prerequisite-free approach, \Model~consistently provides more faithful explanations (i.e., lower \(\text{fid}_{-}\) and \(\text{GEF}\)) than baselines, significantly improving fidelity and robustness.

\textbf{Prerequisite-Satisfied Comparisons.}  
As shown in Table~\ref{tab:ood_comparison}, on the Cora dataset, \Model~achieves up to a 275.7\% improvement over baselines (in concept shift and word domain), with an average improvement of 218.6\% for \(\text{fid}_{+}\), 45.76\% for \(\text{fid}_{-}\), and 65.84\% for \(\text{GEF}\), respectively. Its performance is even comparable to the ideal performance, like PGMExp and CF-GNNExp. On the Motif dataset, \Model~performs well in the size domain, improving \(\text{fid}_{-}\) by 33.42\% and \(\text{GEF}\) by 23.97\%, though its \(\text{fid}_{+}\) performance is lower. In the basis domain, \Model~performs similarly to baselines, further supporting our findings in prerequisite-free comparisons. To summarize, in most cases that align with the SCM used by \Model, it even achieves ideal performance, demonstrating its effectiveness in capturing a more complete GNN decision logic. By adopting a prerequisite-free method, \Model~maintains consistent performance across both prerequisite-free and prerequisite-satisfied scenarios, showcasing greater robustness than baselines.

\begin{table}[t]
\centering

\resizebox{\linewidth}{!}{%
\begin{tabular}{lllllll}
\hline \hline
 & \Model & No LAR & No \(\mathcal{L}_\text{CON}\) & No \( \mathcal{L}_\text{MI} \) & No \(\mathcal{L}_\text{RR}\) & No NPAF \\ \hline
\( \text{fid}_{+} \) \(\uparrow\) & 0.0353 & 0.0720 (+104.0\%) & 0.0703 (+99.15\%) & 0.0693 (+96.32\%) & 0.0720 (+104.0\%) & 0.1430 (+305.1\%) \\
\( \text{fid}_{-} \) \(\downarrow\) & 0.2343 & 0.2970 (-26.76\%) & 0.3033 (-29.45\%) & 0.5050 (-115.5\%) & 0.2767 (-14.21\%) & 0.6503 (-177.6\%) \\
\( \text{GEF} \) \(\downarrow\) & 0.1325 & 0.1345 (-1.509\%) & 0.1329 (-0.302\%) & 0.1052 (+20.60\%) & 0.1374 (-3.698\%) & 0.1948 (-47.02\%) \\ \hline \hline
\end{tabular}%
}
\caption{Ablation study on \Model~various modules.}
\label{tab:ablation}
\end{table}

\subsection{Quality and Efficiency Comparisons (RQ2)} 
Table~\ref{tab:covariate_info} lists the statistic results for the covariate shift scenarios. On complex datasets like Cora, \Model~reduces node density by up to 16.82\% and edge density by 29.23\% compared to PGExp, while also achieving an average speedup of 63.78 times. On simpler datasets like Motif, \Model~delivers performance comparable to baseline methods. Since \Model~provides more faithful explanations than baselines in most cases, these results suggest that \Model~better evaluates the contributions of graph structures to GNN predictions. This demonstrates that, compared to baselines, \Model~captures a more complete GNN decision logic while maintaining similar time complexity, and delivers higher-quality explanations on complex datasets. This demonstrates \Model's scalability and superior generalizability in critical applications compared to baselines.

\subsection{Ablation Study (RQ3)}
Since GNN predictions in OOD scenarios involve uncertainties, accurately evaluating the role and contribution of each module in \Model~becomes challenging. Therefore, we conduct an ablation study on the basis domain of the Motif dataset under in-distribution conditions. We deactivate specific modules by setting the weights of the corresponding modules to zero. These modules include: last action rewards (LAR), the contrastive learning module (\(\mathcal{L}_\text{CON}\)), the reconstruction loss (\(\mathcal{L}_\text{MI}\)), and the reconstruction regularization loss (\(\mathcal{L}_\text{RR}\)). To evaluate the contribution of NPAF, we create a variant by reducing the number of environments to \( K = 1\), meaning that the dataset's sample space is not partitioned.

The results in Table~\ref{tab:ablation} highlight the necessity of each module for optimal performance. When all modules are active, \Model~achieves balanced and robust performance across all metrics. 
Disabling NPAF results in a 177.6\% decrease in \( \text{fid}_{-} \) and a 47.02\% decrease in \(\text{GEF}\), highlighting its critical role in identifying diverse distributions in the dataset's sample space. This further demonstrates that partitioning the sample space enhances \Model's capability to learn differences in GNN decision logic across various distributions. \(\mathcal{L}_\text{MI}\) is the second most influential module, as disabling it reduces \( \text{fid}_{-} \) by 115.5\% but also increases \( \text{fid}_{+} \) and \(\text{GEF}\). This suggests that while it effectively selects important structures, it also incorporates some irrelevant structures into the explanations. Disabling either \(\mathcal{L}_\text{CON}\) or LAR reduces \( \text{fid}_{-} \), indicating that both modules help \Model~capture distribution differences and identify structures relevant to GNN predictions. Disabling \(\mathcal{L}_\text{RR}\) slightly reduces \( \text{fid}_{-} \) and \(\text{GEF}\), suggesting that \(\mathcal{L}_\text{MI}\) alone is insufficient to extract all critical structures for prediction.

\section{Conclusion}
\Model~has represented a major breakthrough in the field of XGNN by uncovering the nearly complete decision logic of GNNs. This research has introduced two key modules: NPAF and GVAG. These modules have collaboratively explored the decision logic of GNNs across the entire dataset's sample space, enhancing the \Model's adaptability. Notably, GVAG's approach to generating explanation subgraphs has eliminated prerequisites on GNN internal accessibility and dataset properties, significantly extending \Model's practical utility. Extensive evaluations across various datasets have confirmed the \Model's capability to provide precise and reliable explanations, underscoring its relevance in real-world applications. In future work, we aim to address the current limitations of \Model, including the inability to handle multi-level OOD issues and the challenge of verifying whether the complete decision logic of GNNs across all distributions has been fully captured, in order to develop a truly comprehensive GNN explainer.

\section*{Acknowledgements}
This research is supported by the ARC Discovery Project DP230100676, Australia, the National Research Foundation, Singapore and Infocomm Media Development Authority under its Trust Tech Funding Initiative. Any opinions, findings and conclusions or recommendations expressed in this material are those of the author(s) and do not reflect the views of the National Research Foundation, Singapore and Infocomm Media Development Authority.

\bibliographystyle{named}
\bibliography{ijcai25}

\begin{thebibliography}{}

\bibitem[\protect\citeauthoryear{Agarwal \bgroup \em et al.\egroup
  }{2023}]{agarwal2023evaluating}
Chirag Agarwal, Owen Queen, Himabindu Lakkaraju, and Marinka Zitnik.
\newblock Evaluating explainability for graph neural networks.
\newblock {\em Scientific Data}, 10(1):144, 2023.

\bibitem[\protect\citeauthoryear{Ahuja \bgroup \em et al.\egroup
  }{2021}]{invariance_pmiboodg_2021}
Kartik Ahuja, Ethan Caballero, Dinghuai Zhang, Jean-Christophe Gagnon-Audet,
  Yoshua Bengio, Ioannis Mitliagkas, and Irina Rish.
\newblock Invariance principle meets information bottleneck for
  out-of-distribution generalization.
\newblock In {\em NeurIPS}, volume~34, pages 3438--3450, 2021.

\bibitem[\protect\citeauthoryear{Amara \bgroup \em et al.\egroup
  }{2022}]{graphframex_tseemgnn_2022}
Kenza Amara, Zhitao Ying, Zitao Zhang, Zhihao Han, Yinan Shan, Ulrik Brandes,
  and Sebastian Schemm.
\newblock Graphframex: Towards systematic evaluation of explainability methods
  for graph neural networks.
\newblock In {\em NeurIPS Workshop}, 2022.

\bibitem[\protect\citeauthoryear{Chen \bgroup \em et al.\egroup
  }{2022}]{learning_ciroodgg_2022}
Yongqiang Chen, Yonggang Zhang, Yatao Bian, Han Yang, MA~Kaili, Binghui Xie,
  Tongliang Liu, Bo~Han, and James Cheng.
\newblock Learning causally invariant representations for out-of-distribution
  generalization on graphs.
\newblock In {\em NeurIPS}, volume~35, pages 22131--22148, 2022.

\bibitem[\protect\citeauthoryear{Chen \bgroup \em et al.\egroup
  }{2023}]{d4explainer_idgnneddd_2023}
Jialin Chen, Shirley Wu, Abhijit Gupta, and Rex Ying.
\newblock D4explainer: in-distribution gnn explanations via discrete denoising
  diffusion.
\newblock In {\em NeurIPS}, pages 78964--78986, 2023.

\bibitem[\protect\citeauthoryear{Chen \bgroup \em et al.\egroup
  }{2024a}]{cheninterpretable}
Yongqiang Chen, Yatao Bian, Bo~Han, and James Cheng.
\newblock How interpretable are interpretable graph neural networks?
\newblock In {\em ICML}, 2024.

\bibitem[\protect\citeauthoryear{Chen \bgroup \em et al.\egroup
  }{2024b}]{chengenerating}
Zhuomin Chen, Jiaxing Zhang, Jingchao Ni, Xiaoting Li, Yuchen Bian, Md~Mezbahul
  Islam, Ananda Mondal, Hua Wei, and Dongsheng Luo.
\newblock Generating in-distribution proxy graphs for explaining graph neural
  networks.
\newblock In {\em ICML}, 2024.

\bibitem[\protect\citeauthoryear{Ding \bgroup \em et al.\egroup
  }{2025}]{ding2025few}
Pengfei Ding, Yan Wang, Guanfeng Liu, Nan Wang, and Xiaofang Zhou.
\newblock Few-shot causal representation learning for out-of-distribution
  generalization on heterogeneous graphs.
\newblock {\em TKDE}, 37(4):1804--1818, 2025.

\bibitem[\protect\citeauthoryear{Fang \bgroup \em et al.\egroup
  }{2024a}]{fang2024evaluating}
Junfeng Fang, Wei Liu, Yuan Gao, Zemin Liu, An~Zhang, Xiang Wang, and Xiangnan
  He.
\newblock Evaluating post-hoc explanations for graph neural networks via
  robustness analysis.
\newblock In {\em NeurIPS}, volume~36, 2024.

\bibitem[\protect\citeauthoryear{Fang \bgroup \em et al.\egroup
  }{2024b}]{fang2024regularization}
Junfeng Fang, Guibin Zhang, Kun Wang, Wenjie Du, Yifan Duan, Yuankai Wu, Roger
  Zimmermann, Xiaowen Chu, and Yuxuan Liang.
\newblock On regularization for explaining graph neural networks: An
  information theory perspective.
\newblock {\em TKDE}, 2024.

\bibitem[\protect\citeauthoryear{Fey and Lenssen}{2019}]{FeyLenssen2019}
Matthias Fey and Jan~E. Lenssen.
\newblock Fast graph representation learning with {PyTorch Geometric}.
\newblock In {\em ICLR Workshop}, 2019.

\bibitem[\protect\citeauthoryear{Golmaei and Luo}{2021}]{golmaei2021deepnote}
Sara~Nouri Golmaei and Xiao Luo.
\newblock Deepnote-gnn: predicting hospital readmission using clinical notes
  and patient network.
\newblock In {\em ACM-BCB}, pages 1--9, 2021.

\bibitem[\protect\citeauthoryear{Gui \bgroup \em et al.\egroup
  }{2022}]{good_goodb_2022}
Shurui Gui, Xiner Li, Limei Wang, and Shuiwang Ji.
\newblock Good: A graph out-of-distribution benchmark.
\newblock In {\em NeurIPS}, volume~35, pages 2059--2073, 2022.

\bibitem[\protect\citeauthoryear{Kawamoto \bgroup \em et al.\egroup
  }{2018}]{mean_ftgnngp_2018}
Tatsuro Kawamoto, Masashi Tsubaki, and Tomoyuki Obuchi.
\newblock Mean-field theory of graph neural networks in graph partitioning.
\newblock In {\em NeurIPS}, volume~31, 2018.

\bibitem[\protect\citeauthoryear{Kingma and Welling}{2014}]{kingma2013auto}
Diederik~P Kingma and Max Welling.
\newblock Auto-encoding variational bayes.
\newblock {\em stat}, 1050:1, 2014.

\bibitem[\protect\citeauthoryear{Koch \bgroup \em et al.\egroup
  }{2022}]{koch2022hidden}
Lisa~M Koch, Christian~M Sch{\"u}rch, Arthur Gretton, and Philipp Berens.
\newblock Hidden in plain sight: Subgroup shifts escape ood detection.
\newblock In {\em Machine Learning Research}, volume 172, pages 726--740. PMLR,
  2022.

\bibitem[\protect\citeauthoryear{Koch \bgroup \em et al.\egroup
  }{2024}]{koch2024distribution}
Lisa~M Koch, Christian~F Baumgartner, and Philipp Berens.
\newblock Distribution shift detection for the postmarket surveillance of
  medical ai algorithms: a retrospective simulation study.
\newblock {\em NPJ Digital Medicine}, 7(1):120, 2024.

\bibitem[\protect\citeauthoryear{Kubo and Difallah}{2024}]{kubo2024xgexplainer}
Ryoji Kubo and Djellel Difallah.
\newblock Xgexplainer: Robust evaluation-based explanation for graph neural
  networks.
\newblock In {\em SDM}, pages 64--72. SIAM, 2024.

\bibitem[\protect\citeauthoryear{Li \bgroup \em et al.\egroup
  }{2022}]{li2022graph}
Michelle~M Li, Kexin Huang, and Marinka Zitnik.
\newblock Graph representation learning in biomedicine and healthcare.
\newblock {\em Nature Biomedical Engineering}, 6(12):1353--1369, 2022.

\bibitem[\protect\citeauthoryear{Liu \bgroup \em et al.\egroup
  }{2021}]{JMLRv22210343}
Meng Liu, Youzhi Luo, Limei Wang, Yaochen Xie, Hao Yuan, Shurui Gui, Haiyang
  Yu, Zhao Xu, Jingtun Zhang, Yi~Liu, Keqiang Yan, Haoran Liu, Cong Fu, Bora~M
  Oztekin, Xuan Zhang, and Shuiwang Ji.
\newblock {DIG}: A turnkey library for diving into graph deep learning
  research.
\newblock {\em Journal of Machine Learning Research}, 22(240):1--9, 2021.

\bibitem[\protect\citeauthoryear{Lucic \bgroup \em et al.\egroup
  }{2022}]{cf_cegnn_2022}
Ana Lucic, Maartje~A Ter~Hoeve, Gabriele Tolomei, Maarten De~Rijke, and
  Fabrizio Silvestri.
\newblock Cf-gnnexplainer: Counterfactual explanations for graph neural
  networks.
\newblock In {\em AISTATS}, pages 4499--4511. PMLR, 2022.

\bibitem[\protect\citeauthoryear{Luo \bgroup \em et al.\egroup
  }{2020}]{parameterized_egnn_2020}
Dongsheng Luo, Wei Cheng, Dongkuan Xu, Wenchao Yu, Bo~Zong, Haifeng Chen, and
  Xiang Zhang.
\newblock Parameterized explainer for graph neural network.
\newblock In {\em NeurIPS}, volume~33, pages 19620--19631. Curran Associates,
  Inc., 2020.

\bibitem[\protect\citeauthoryear{Miller \bgroup \em et al.\egroup
  }{2020}]{miller2020adversarial}
David~J Miller, Zhen Xiang, and George Kesidis.
\newblock Adversarial learning targeting deep neural network classification: A
  comprehensive review of defenses against attacks.
\newblock In {\em IEEE}, volume 108, pages 402--433. IEEE, 2020.

\bibitem[\protect\citeauthoryear{Min \bgroup \em et al.\egroup
  }{2024}]{min2024graph}
Xin Min, Wei Li, Ruiqi Han, Tianlong Ji, and Weidong Xie.
\newblock Graph neural collaborative filtering with medical content-aware
  pre-training for treatment pattern recommendation.
\newblock {\em Pattern Recognition Letters}, 185:210--217, 2024.

\bibitem[\protect\citeauthoryear{Pope \bgroup \em et al.\egroup
  }{2019}]{explainability_mgcnn_2019}
Phillip~E. Pope, Soheil Kolouri, Mohammad Rostami, Charles~E. Martin, and Heiko
  Hoffmann.
\newblock Explainability methods for graph convolutional neural networks.
\newblock In {\em IEEE/CVF CVPR}, pages 10772--10781, 2019.

\bibitem[\protect\citeauthoryear{Qiu \bgroup \em et al.\egroup
  }{2024}]{qiu2024generating}
Dazhuo Qiu, Mengying Wang, Arijit Khan, and Yinghui Wu.
\newblock Generating robust counterfactual witnesses for graph neural networks.
\newblock {\em arXiv preprint arXiv:2404.19519}, 2024.

\bibitem[\protect\citeauthoryear{Simonovsky and
  Komodakis}{2018}]{simonovsky2018graphvae}
Martin Simonovsky and Nikos Komodakis.
\newblock Graphvae: Towards generation of small graphs using variational
  autoencoders.
\newblock In {\em ICANN}, pages 412--422. Springer, 2018.

\bibitem[\protect\citeauthoryear{Togninalli \bgroup \em et al.\egroup
  }{2019}]{wasserstein_wlgk_2019}
Matteo Togninalli, Elisabetta Ghisu, Felipe Llinares-L{\'o}pez, Bastian Rieck,
  and Karsten Borgwardt.
\newblock Wasserstein weisfeiler-lehman graph kernels.
\newblock In {\em NeurIPS}, volume~32, 2019.

\bibitem[\protect\citeauthoryear{Veli{\v{c}}kovi{\'c} \bgroup \em et al.\egroup
  }{2018}]{gat_2018}
Petar Veli{\v{c}}kovi{\'c}, Guillem Cucurull, Arantxa Casanova, Adriana Romero,
  Pietro Lio, and Yoshua Bengio.
\newblock Graph attention networks.
\newblock {\em stat}, 1050(20):10--48550, 2018.

\bibitem[\protect\citeauthoryear{Vu and Thai}{2020}]{pgme_pgmegnn_2020}
Minh Vu and My~T Thai.
\newblock Pgm-explainer: Probabilistic graphical model explanations for graph
  neural networks.
\newblock In {\em NeurIPS}, volume~33, pages 12225--12235, 2020.

\bibitem[\protect\citeauthoryear{Wang and Shen}{2023}]{wanggnninterpreter}
Xiaoqi Wang and Han~Wei Shen.
\newblock Gnninterpreter: A probabilistic generative model-level explanation
  for graph neural networks.
\newblock In {\em ICLR}, 2023.

\bibitem[\protect\citeauthoryear{Wu \bgroup \em et al.\egroup
  }{2021}]{discovering_irgnn_2022}
Yingxin Wu, Xiang Wang, An~Zhang, Xiangnan He, and Tat-Seng Chua.
\newblock Discovering invariant rationales for graph neural networks.
\newblock In {\em ICLR}, 2021.

\bibitem[\protect\citeauthoryear{Xiong \bgroup \em et al.\egroup
  }{2021}]{xiong2021heterogeneous}
Kai Xiong, Xiao Ding, Li~Du, Ting Liu, and Bing Qin.
\newblock Heterogeneous graph knowledge enhanced stock market prediction.
\newblock {\em AI Open}, 2:168--174, 2021.

\bibitem[\protect\citeauthoryear{Xu \bgroup \em et al.\egroup
  }{2024}]{xu2024adaptive}
Rongwei Xu, Guanfeng Liu, Yan Wang, Xuyun Zhang, Kai Zheng, and Xiaofang Zhou.
\newblock Adaptive hypergraph network for trust prediction.
\newblock In {\em ICDE}, pages 2986--2999. IEEE, 2024.

\bibitem[\protect\citeauthoryear{Ying \bgroup \em et al.\egroup
  }{2019}]{gnnexplainer_gegnn_2019}
Zhitao Ying, Dylan Bourgeois, Jiaxuan You, Marinka Zitnik, and Jure Leskovec.
\newblock Gnnexplainer: Generating explanations for graph neural networks.
\newblock In {\em NeurIPS}, volume~32, pages 9244--9255. Curran Associates,
  Inc., 2019.

\bibitem[\protect\citeauthoryear{Yuan \bgroup \em et al.\egroup
  }{2020}]{xgnn_tmegnn_2020}
Hao Yuan, Jiliang Tang, Xia Hu, and Shuiwang Ji.
\newblock Xgnn: Towards model-level explanations of graph neural networks.
\newblock In {\em ACM KDD}, pages 430--438, 2020.

\bibitem[\protect\citeauthoryear{Zhang \bgroup \em et al.\egroup
  }{2022}]{rgnnsm_2022}
Wenjun Zhang, Zhensong Chen, Jianyu Miao, and Xueyong Liu.
\newblock Research on graph neural network in stock market.
\newblock {\em Procedia Computer Science}, 214:786--792, 2022.

\bibitem[\protect\citeauthoryear{Zhang \bgroup \em et al.\egroup
  }{2023}]{mixupexplainer_gegnnda_2023}
Jiaxing Zhang, Dongsheng Luo, and Hua Wei.
\newblock Mixupexplainer: Generalizing explanations for graph neural networks
  with data augmentation.
\newblock In {\em ACM KDD}, pages 3286--3296, 2023.

\bibitem[\protect\citeauthoryear{Zhu \bgroup \em et al.\egroup
  }{2023}]{zhu2023domain}
Jiajie Zhu, Yan Wang, Feng Zhu, and Zhu Sun.
\newblock Domain disentanglement with interpolative data augmentation for
  dual-target cross-domain recommendation.
\newblock In {\em RecSys}, pages 515--527, 2023.

\end{thebibliography}

\clearpage
\appendix
\section{Supplementary Material on Methodology}
\subsection{Symbol Summary}\label{app:symbol}
Table~\ref{tab:symbols} provides a summary of all symbols used in this paper, along with their descriptions.
\begin{table}[h]
    \centering
    
    \resizebox{0.95\linewidth}{!}{ % 
        \begin{tabular}{c l}
            \hline \hline
            \textbf{Symbol} & \textbf{Description} \\ 
            \hline
            $\mathcal{V}$ & Node set \\
            $\mathcal{E}$ & Edge set \\
            $\mathcal{X}$ & Node features \\
            $A$ & Adjacency matrix \\
            $n$ & Number of nodes \\
            $m$ & Number of edges \\
            $\mathcal{M}$ & Target GNN model \\
            $Y$ & Graph/node classification label \\
            $E$ & Environment variables \\
            $G_c$ & Explanation subgraph \\
            $G_s$ & Complement graph after excluding the explanation subgraph \\
            $G_\text{test}$ & Graph in testing dataset \\
            $X_{str}$ & Structure-based features for nodes \\
            $H_{str}$ & Structure-based embeddings for nodes \\
            $H_G$ & Structure-based embeddings for graphs \\
            $K$ & Number of potential environments \\
            $env$ & Environmental embedding set\\
            $E_{str}$ & Structure-based environment label set \\
            $E^s_k$ & The $k$-th structure-based environment label \\
            $E_{feat}$ & Feature-based environment label set \\
            $E^f_k$ & The $k$-th feature-based environment label \\
            $\mathcal{V}_c$ & Nodes causally related to the classification label \\
            $\mathcal{V}_s$ & Nodes causally related to the environment \\
            $\mathbf{H}$ & Feature-based node embeddings \\
            $\mathbf{h}_i$ & Feature-based embedding for node $i$ \\
            $\mathbf{h}_G$ & Feature-based embedding for graph $G$ \\
            $\mathbf{e}_i$ & Environmental embedding for node $i$ \\
            $\mathbf{e}_G$ & Environmental embedding for graph $G$ \\
            $\boldsymbol{\mu}_i$ & Mean of the node-invariant representation distribution \\
            & of node $i$ \\
            $\boldsymbol{\mu}_G$ & Mean of the graph-invariant representation distribution of $G$ \\
            $\log (\boldsymbol{\sigma}_i^2)$ & Log-variance of the node-invariant representation \\ 
            & distribution of node $i$ \\
            $\log (\boldsymbol{\sigma}_G^2)$ & Log-variance of the graph-invariant representation \\
            & distribution of graph $G$ \\
            $\mathbf{z}_i$ & Node-invariant representation of node $i$ \\
            $\mathbf{z}_\text{node}$ & Node-invariant embeddings \\
            $\mathbf{z}_G$ & Graph-invariant embedding of graph $G$ \\
            $\boldsymbol{\epsilon}$ & Random noise \\
            $D_{\text{KL}}(\cdot \parallel \cdot)$ & KL divergence \\
            $JS(\cdot \| \cdot)$ & JS divergence \\
            $q_{\phi_{1}}(\mathbf{z}_i|\mathbf{h}_i, env)$ & Distribution modeled by the NodeVAE encoder\\
            $p_{\theta_1}(\mathbf{h}_i|\mathbf{z}_i, env)$ & Distribution modeled by the NodeVAE decoder\\
            $q_{\phi_2}(\mathbf{z}_G|G, env)$ & Distribution modeled by the GVAG encoder\\
            $p_{\theta_2}(v_i | \mathbf{z}_G, \mathbf{z}_i, env)$ & Node existence probability distribution modeled by \\
            & GVAG decoder\\
            $p_{\theta_3}(e_{ij}|\mathbf{z}_G, \mathbf{z}_i, \mathbf{z}_j, env)$ & Edge existence probability distribution modeled by \\
            & GVAG decoder\\
            $p(\mathbf{z})$ & Prior distribution of graph-invariant representations\\
            $\text{Prob}({v_i})$ & Node existence probability of node $i$\\
            $\text{Prob}({e_{ij}})$ & Edge existence probability of edge $e_{ij}$\\
            $\mathcal{L}_\text{MI}$ & Subgraph reconstruction loss/$MI$ loss\\
            $\mathcal{L}_{\text{RR}}$ & Reconstruction regularization loss\\
            $\mathcal{L}_\text{NodeVAE}$ & NodeVAE loss\\
            $\mathcal{R}_\text{causal}$ & Causal structure regularization\\
            $\mathcal{R}_\text{hinge}$ & Hinge regularization\\
            $\mathcal{R}_\text{subg\_node}$ & Subgraph node count regularization\\
            $\mathcal{L}_\text{CON}$ & Contrastive loss\\
            $\text{LAR}$ & Last action rewards\\
            $\text{fid}_{-}$ & Negative Fidelity\\
            $\text{fid}_{+}$ & Positive Fidelity\\
            $\text{GEF}$ & Unfaithfulness\\
            $\rho_v$ & Node density relative to the original graph\\
            $\rho_e$ & Edge density relative to the original graph\\
            $T$ & Response time (second)\\
            \hline \hline
        \end{tabular}
    }
\caption{List of symbols and their descriptions.}
\label{tab:symbols}
\end{table}

\subsection{Subgraph Reconstruction Algorithm}\label{app:recon_algor}

\begin{algorithm}[]
\small
\caption{Sample-based subgraph reconstruction algorithm}
\label{alg:training_genereate}
\begin{algorithmic}[1]
\Require \begin{tabular}[t]{@{}l}
    Input parameters: \\
    \textit{edge\_index}: Edge set of original graph \\
    \textit{node\_prob}: Node existence probability \\
    \textit{link\_prob}: Edge existence probability \\
    \textit{max\_nodes}: Maximum number of nodes in subgraph \\
    \textit{start\_nid}: Start node of generating subgraph \\
    \textit{density}: Subgraph density limit \\
    \textit{max\_iter}: Maximum number of iterations \\
    \textit{min\_edges}: Minimum number of edges in subgraph
\end{tabular}

% \textit{edge\_index, node\_prob, link\_prob, max\_nodes, start\_nid, density, max\_iter, min\_edges}

\State \textbf{Initialize:}
\State Ensure \textit{max\_nodes} does not exceed the number of nodes available in the original graph
\State Adjust \textit{min\_edges} based on the \textit{density}

\State Add a small epsilon to \textit{node\_prob} and \textit{link\_prob} to prevent computational issues 

% \Comment{Handle zero probabilities for nodes and links}

\State Initialize node selection vector \textit{current\_node}
\Statex \textbf{Sampling nodes for the subgraph:}

\If{\textit{start\_nid} is provided}
    \State Set the corresponding index in \textit{current\_node} to True
\EndIf
\State Copy \textit{node\_prob} as sampling probabilities \(Prob_n\) 
\For{\textit{iter} from 1 to \textit{max\_iter}} 
% \Comment{Node Sampling Loop (up to max\_iter iterations):}
    \State For nodes in \textit{current\_node}, set \(Prob_n\) to -1
    \State Sample new nodes based on \(Prob_n\) and update \textit{current\_node}
    \If{sum(\textit{current\_node}) $\geq$ \textit{max\_nodes}}
        \State \textbf{break}
    \EndIf
\EndFor

\If{sum(\textit{current\_node}) $<$ \textit{max\_nodes}} 
    \State For nodes in \textit{current\_node}, set \(Prob_n\) to -1
% \Comment{Adjust node selection if necessary:}
    \State Select new nodes based on \(Prob_n\) and update \textit{current\_node}
\Else
    \State Prune nodes from \textit{current\_node} based on \textit{node\_prob} to fit within \textit{max\_nodes}
\EndIf

\State Reset \(Prob_n\) to \textit{node\_prob}
\State For nodes in \textit{current\_node}, set \(Prob_n\) to 1
\Statex \textbf{Sampling edges for the subgraph based on the nodes within the subgraph:}
\State Initialize edge selection vector \textit{current\_edge}
\For{\textit{iter} from 1 to \textit{max\_iter}} 
    \State Copy \textit{link\_prob} as sampling probabilities \(Prob_e\)
    \State For links in \textit{current\_edge}, set \(Prob_e\) to 0
    \State Recompute \(Prob_e\) using \(Prob_n\) and \(Prob_e\)
% \Comment{Edge Sampling Loop (up to max\_iter iterations):}
    \State Sample links based on \(Prob_e\)
    \State Update \textit{current\_link} based on sampled links
    \If{edge density $>$ \textit{density}}
        \State \textbf{break}
    \EndIf
\EndFor

\State If node has no edge, remove it from \textit{current\_node}
\State Calculate total graph probability \textit{total\_graph\_prob} based on \textit{node\_prob} and \textit{link\_prob}
\Statex
\State \Return \textit{current\_node, current\_link, total\_graph\_prob}
\end{algorithmic}
\end{algorithm}

\begin{algorithm}[t]
\small
\caption{Edge first reconstruction algorithm}
\label{alg:eval_graph_recon}
\begin{algorithmic}[1]
\Require \begin{tabular}[t]{@{}l}
    Input parameters: \\
    \textit{edge\_index}: Edge set of original graph \\
    \textit{node\_prob}: Node existence probability \\
    \textit{link\_prob}: Edge existence probability \\
    \textit{max\_nodes}: Maximum number of nodes in subgraph \\
    \textit{start\_nid}: Start node of generating subgraph \\
    \textit{density}: Subgraph density limit \\
    % \textit{max\_iter}: Maximum number of iterations \\
    \textit{min\_edges}: Minimum number of edges in subgraph
\end{tabular}

\Statex \textbf{Initialize:}
\State Initialize node selection vector \textit{current\_node}
\State Initialize edge selection vector \textit{current\_edge}
\State Adjust \textit{min\_edges} based on \textit{density}
\State Add a small epsilon to \textit{node\_prob} and \textit{link\_prob} to prevent computational issues

% \Comment{Handle zero probabilities for nodes and links}
\Statex \textbf{Prepare edge existence probability:}
\If{\textit{start\_nid} is provided} 
% \Comment{Set Start Node:}
    \State Set the corresponding index in \textit{current\_node} to True
\EndIf

% \State Create \textit{current\_link} as a zero vector of boolean type \Comment{Initialize link selection:}

\State \textit{total\_graph\_prob} $\gets$ \textit{0.0}

\State \textit{current\_node\_prob} $\gets$ \textit{node\_prob}
\State \textit{current\_link\_prob} $\gets$ \textit{link\_prob}
\State Recompute \textit{current\_link\_prob} using \textit{current\_node\_prob} and \textit{current\_link\_prob}
% \State \textit{current\_link\_prob} $\gets$ recompute edge probability based on \textit{current\_node} and \textit{current\_link}

% \State \textbf{Select Edges:}
\State \textit{max\_edges} $\gets$ \textit{ceil(density * \(|edge\_index|\))}
\If{\textit{max\_edges} $\leq$ \textit{min\_edges}}
    \State \textit{max\_edges} $\gets$ \textit{min\_edges}
\EndIf
% \State \textit{max\_edges} $\gets$ \textit{max\_edges} if \textit{max\_edges} $\geq$ \textit{min\_edges} else \textit{min\_edges}
\Statex \textbf{Select edges for the explanatory subgraph based on their probabilities:}
\State \textit{sorted\_edge\_prob, sorted\_eid} $\gets$ \textit{topk(current\_link\_prob, k=max\_edges)}
\State For edge id in \textit{sorted\_eid}, set \textit{current\_edge} to True
\State Calculate total graph probability \textit{total\_graph\_prob} based on \textit{sorted\_edge\_prob}
\Statex \textbf{Update Nodes Based on Selected Edges:}
\State Get source nodes of edges in \textit{current\_edge}, as \textit{src\_nodes} 
\State Get destination nodes of edges in \textit{current\_edge}, as \textit{dst\_nodes}
\State For nodes in \textit{src\_nodes} or \textit{dst\_nodes}, set \textit{current\_node} to True
\Statex
\State \Return \textit{current\_node, current\_link, total\_graph\_prob}
\end{algorithmic}
\end{algorithm}

Algorithm~\ref{alg:training_genereate} presents the pseudocode for generating subgraphs during the training phase. In essence, during the training phase, we first sample nodes that will appear on the explanation subgraph based on their existence probability, assigning these sampled nodes a probability of $1.0$. Then, we recalculate the existence probabilities for each edge by integrating both the node existence probability and the edge existence probability relevant to the current subgraph. Next, these recalculated probabilities are subsequently utilized to sample edges that will appear on the explanation subgraph. This methodology ensures that the probabilities of both nodes and edges are considered concurrently in generating the explanation subgraph, which can help in maintaining connectivity within the subgraph.

Moreover, through this random sampling process, GVAG effectively explores the entire space of potential subgraphs and avoids focusing on a limited set of nodes or edges, thereby enhancing the robustness and generalization of the generated explanation.

In the testing phase, GVAG directly uses the node existence probability and edge existence probability to calculate the final probability of each edge appearing on the explanation subgraph, and add these edges to the explanation subgraph according to the probability. The corresponding pseudocode is presented in Algorithm~\ref{alg:eval_graph_recon}.

\subsection{Computational Complexity Analysis of Subgraph Generation During Training Phase}
We adopt a sampling-based subgraph generation algorithm, detailed in Algorithm~\ref{alg:training_genereate}, with its complexity influenced by various operations. The initialization steps, including adjustments to \texttt{max\_nodes} and \texttt{min\_edges}, are constant operations with a complexity of \( O(1) \), while adding \(\epsilon\) to zero probabilities involves linear operations, resulting in a combined complexity of \( O(n) + O(m) \). The node sampling loop iterates up to \texttt{max\_iter} times, with each iteration involving probability calculations and updates for all nodes, leading to a complexity of \( O(n \cdot \text{max\_iter}) \). Similarly, the edge sampling loop processes all edges within each iteration, requiring recomputation of probabilities and sampling, contributing a complexity of \( O(m \cdot \text{max\_iter}) \). Post-processing adjustments, such as sorting and selecting top \( k \) elements for nodes and edges, add logarithmic factors, typically \( O(n \log n) \) and \( O(m \log m) \), respectively. Considering \( n > m \), \( \text{max\_iter} > \log n \), and \( \text{max\_iter} > \log m \), the overall time complexity of the algorithm is \( O(n \cdot \text{max\_iter}) \).

\subsection{Computation Complexity Analysis of Subgraph Generation During Evaluation Phase}
The computational complexity of the Algorithm~\ref{alg:eval_graph_recon} primarily depends on the operations performed on nodes and edges within the graph. Initially, adjusting the probability vectors for zero probabilities, which are operations linear in terms of the number of nodes \(n\) and edges \(m\), contributes a complexity of \(O(n + m)\). This setup is followed by the critical step of selecting edges based on updated probabilities, involving a sorting operation. Since the edges are sorted to select the top edges based on their probability, this step incurs a complexity of \(O(m \log m)\), which is the most computationally intensive part of the function. Updating the node selection based on the edges selected is relatively straightforward and operates linearly with respect to the number of selected edges, hence contributing an additional linear term. Overall, the sorting of edge probabilities dominates the computational complexity, making the function's total complexity mainly governed by \(O(m \log m)\) with an additional linear component due to initialization and node updates based on selected edges.

\section{Supplementary Material on Experiments}\label{app:exp_sec}
\subsection{Experimental Setup}\label{app:exp_setup}
Our experimental were conducted on an AMD EPYC 9754 CPU, an NVIDIA 4090D GPU with 24GB G6X memory, and 60GB of RAM. The software environment includes Python 3.10, CUDA 12.1, and PyTorch 2.1.0. 

\textbf{The Hyper-Parameter Settings.} 
The key hyper-parameter settings for training and evaluation are shown in Table~\ref{tab:hyperparam}. These hyper-parameters are carefully selected based on prior empirical results and tuning experiments to balance model fidelity and efficiency.

\begin{table}[]
\centering

\resizebox{0.9\columnwidth}{!}{%
\begin{tabular}{lll}
\hline \hline
Hyper-parameters & Cora & Motif \\ \hline
Learning Rate & 0.01 & 0.005 \\
Weight Decay & 1.00E-04 & 1.00E-04 \\
Structure Infer Epochs & 5 & 5 \\
Number Environments $K$ & 4 & 5 \\
Number Epochs & 1 & 10 \\
Batch Size & 64 & 64 \\
Prior Subgraph Max Nodes & 60 & 7 \\
Prior Subgraph Min Nodes & 15 & 5 \\
Prior Subgraph Density & 0.35 & 0.1 \\
Recon Loss Weight \(\omega_\text{RECON}\) & 2 & 2 \\
Contrastive Loss Weight \(\omega_\text{CON}\) & 0.5 & 0.5 \\
Last Action Rewards Weight \(\omega_\text{LAR}\) & 1 & 1 \\ \hline \hline
\end{tabular}%
}
\caption{Hyper-parameters settings.}
\label{tab:hyperparam}
\end{table}

\subsection{Analysis of Hyper-Parameter Sensitivity}\label{app:hyperparam_sensiti}
We also test the impact of several hyper-parameters on the quality of the final generated explanation subgraph, including edge density, last action rewards weight and reconstruction weight. 

\begin{figure}[ht]
    \centering
    \includegraphics[width=\linewidth]{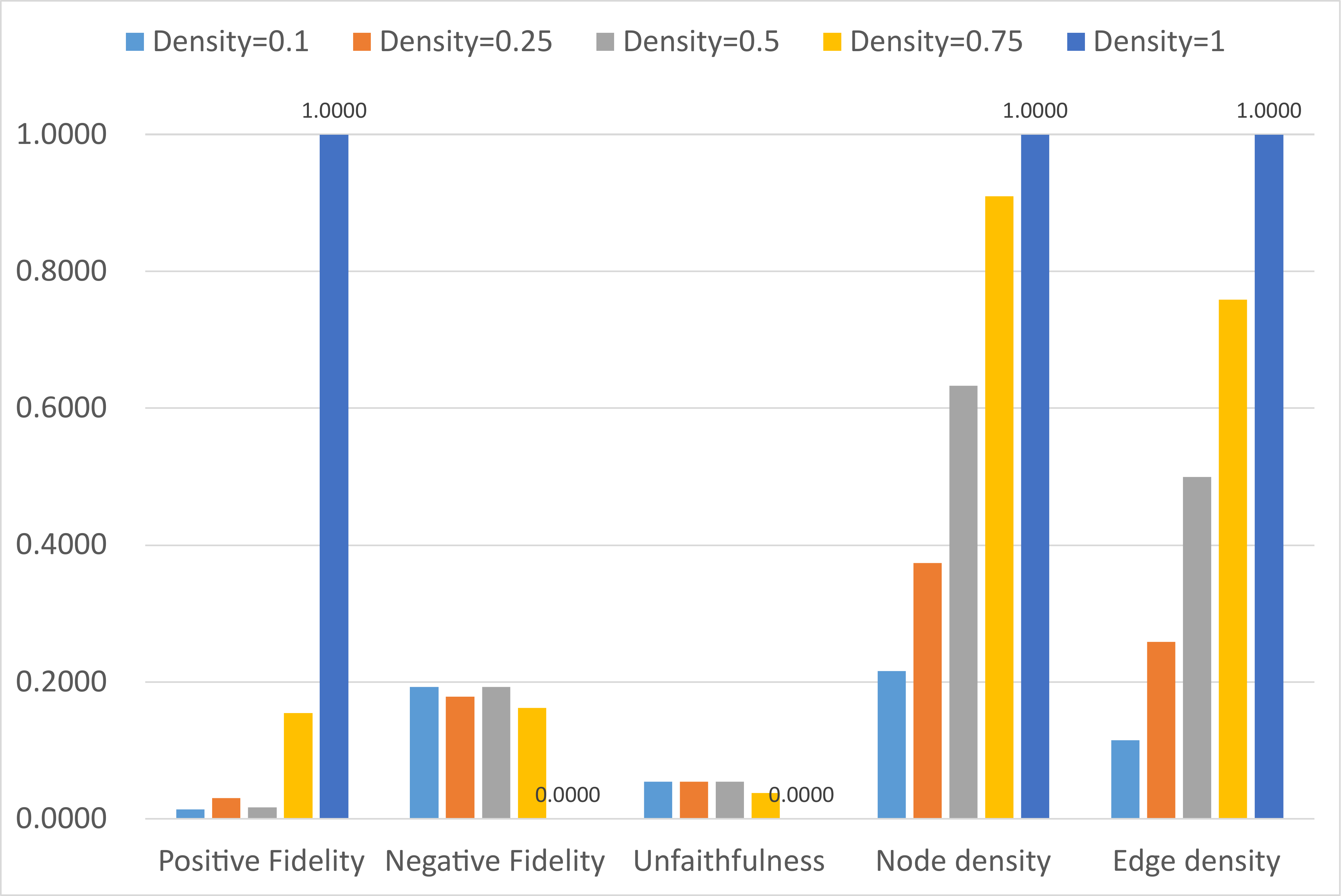}
    \caption{Hyper-parameter sensitivity study on different edge densities.}
    \label{fig:diff_density}
\end{figure}

\paragraph{Different Edge Density.}This study examines how edge density impacts the model's ability to generate effective explanations, with results summarized in Figure~\ref{fig:diff_density}. As edge density increases from 0.1 to 1.0, a distinct trend in the performance metrics emerges:
\begin{itemize}[leftmargin=*]
    \item \textbf{Positive Fidelity:} Variations in positive fidelity suggest that explanation subgraphs with higher densities include more critical features essential for GNN predictions, thereby enhancing positive fidelity by better aligning with GNN predictions.
    \item \textbf{Negative Fidelity:} Adjustments in negative fidelity with varying densities indicate that lower densities, which result in sparser subgraphs, may omit crucial features necessary for the GNN's decision-making process.
    \item \textbf{Unfaithfulness:} Changes in unfaithfulness show that denser explanation subgraphs are likely to exhibit lower unfaithfulness, implying that denser subgraphs may better align with the predictions of the original graph, thus offering more faithful interpretations.
\end{itemize}
This analysis underscores the importance of managing edge density to balance the trade-offs between comprehensiveness and simplicity in explanation subgraphs. Lower densities, while easier to interpret, might miss critical information; conversely, higher densities, though potentially more complex, provide a more detailed and accurate representation of the factors influencing the model's decisions. Achieving this balance is crucial for ensuring that explanations are both informative and practically useful, therefore meeting the needs of real-world applications.
% Striking this balance is vital for creating explanations that are both informative and practical for real-world applications.

\begin{figure}[t]
    \centering
    \includegraphics[width=\linewidth]{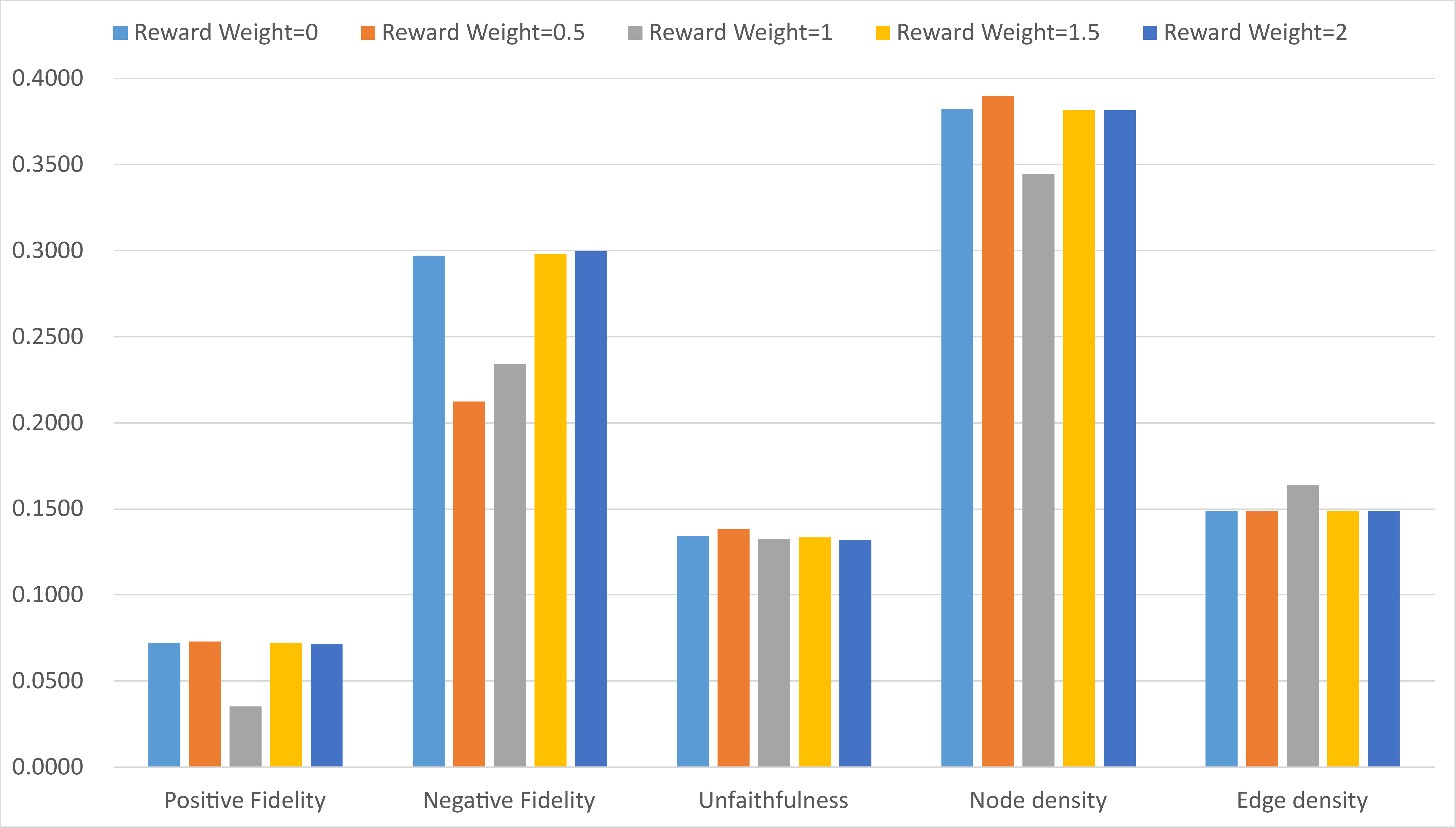}
    \caption{Hyper-parameter sensitivity study on different weights of LAR.}
    \label{fig:diff_reward}
\end{figure}

\paragraph{Different Weights of Last Action Rewards.} 
This study explores how varying the weights of the last action rewards influences \Model's performance, with results presented in Figure~\ref{fig:diff_reward}. 
\begin{itemize}[leftmargin=*]
    \item \textbf{Positive Fidelity:} Positive fidelity remains relatively stable across different reward weights, suggesting that adjustments in the last action rewards do not significantly affect the model's capability to include critical graph structures in the explanation subgraphs.
    \item \textbf{Negative Fidelity:} We observe a regular increase in negative fidelity as the reward weight increases, indicating that excessively high reward weights might overly penalize the model's errors, potentially leading to the exclusion of relevant structures.
    \item \textbf{Unfaithfulness:} Unfaithfulness demonstrates a decreasing trend with higher reward weights, which implies that the explanations become more aligned with the original graph's predictions, enhancing their faithfulness and reliability.
\end{itemize}
These findings indicate that while a higher reward weight can enhance the faithfulness of explanations, it may simultaneously compromise negative fidelity by penalizing the model too harshly. Therefore, it is essential to finely tune the last action rewards to maintain a balance between the depth and accuracy of the explanations produced by \Model, ensuring that they are comprehensive yet precise.

% This balance is critical for generating explanations that are both insightful and practically useful.

\begin{figure}[t]
    \centering
    \includegraphics[width=\linewidth]{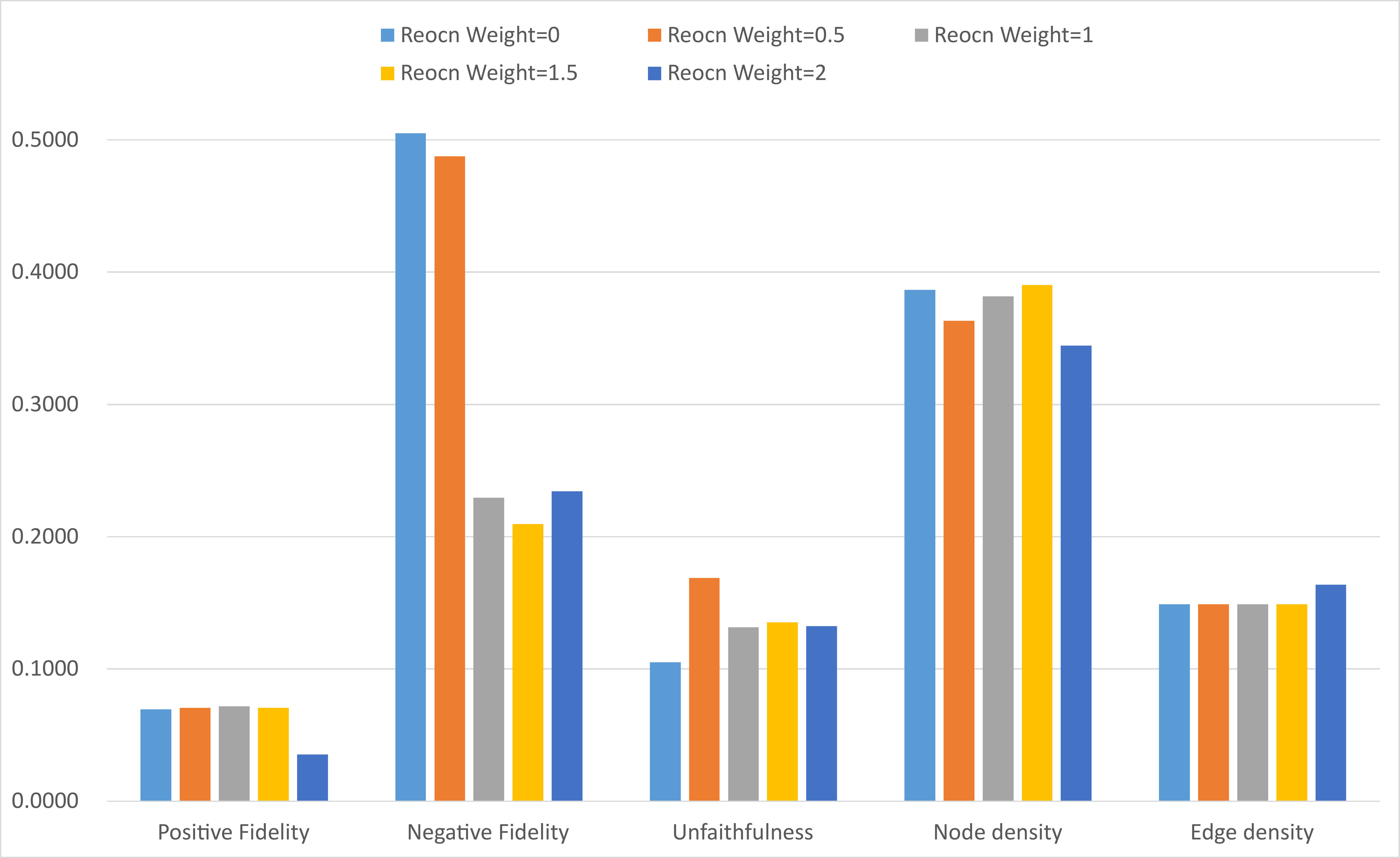}
    \caption{Hyper-parameter sensitivity study on different weights of reconstruction loss.}
    \label{fig:diff_recon}
\end{figure}

\paragraph{Different Weights of Reconstruction Loss.}
This study examines the effect of varying reconstruction loss weights on \Model's performance, with results illustrated in Figure~\ref{fig:diff_recon}. \begin{itemize}[leftmargin=*]
    \item \textbf{Positive Fidelity:} Positive fidelity shows a positive correlation with increasing reconstruction loss weight, climbing from 0 to 2. This trend suggests that higher weights prompt the model to preserve more crucial structures within the explanation subgraphs, thereby boosting positive fidelity.
    \item \textbf{Negative Fidelity:} Conversely, negative fidelity decreases as the reconstruction loss weight increases, highlighting that enhanced penalties for incorrect explanations help the model omit non-essential structures, thus refining the fidelity of its explanations.
    \item \textbf{Unfaithfulness:} Unfaithfulness does not exhibit a consistent trend with changes in reconstruction loss weight, indicating that while reconstruction loss aids in refining explanations, its effect on the faithfulness may be influenced by other factors within the model.
\end{itemize}
These findings illustrate the critical role that reconstruction loss weight plays in explanations generated by \Model. Properly setting this weight can enhances the fidelity and accuracy of explanations. 

\subsection{Explanation Subgraph Examples}\label{app:explanation_subgraphs}
\begin{figure*}[t]
    \centering
    \includegraphics[width=0.85\textwidth]{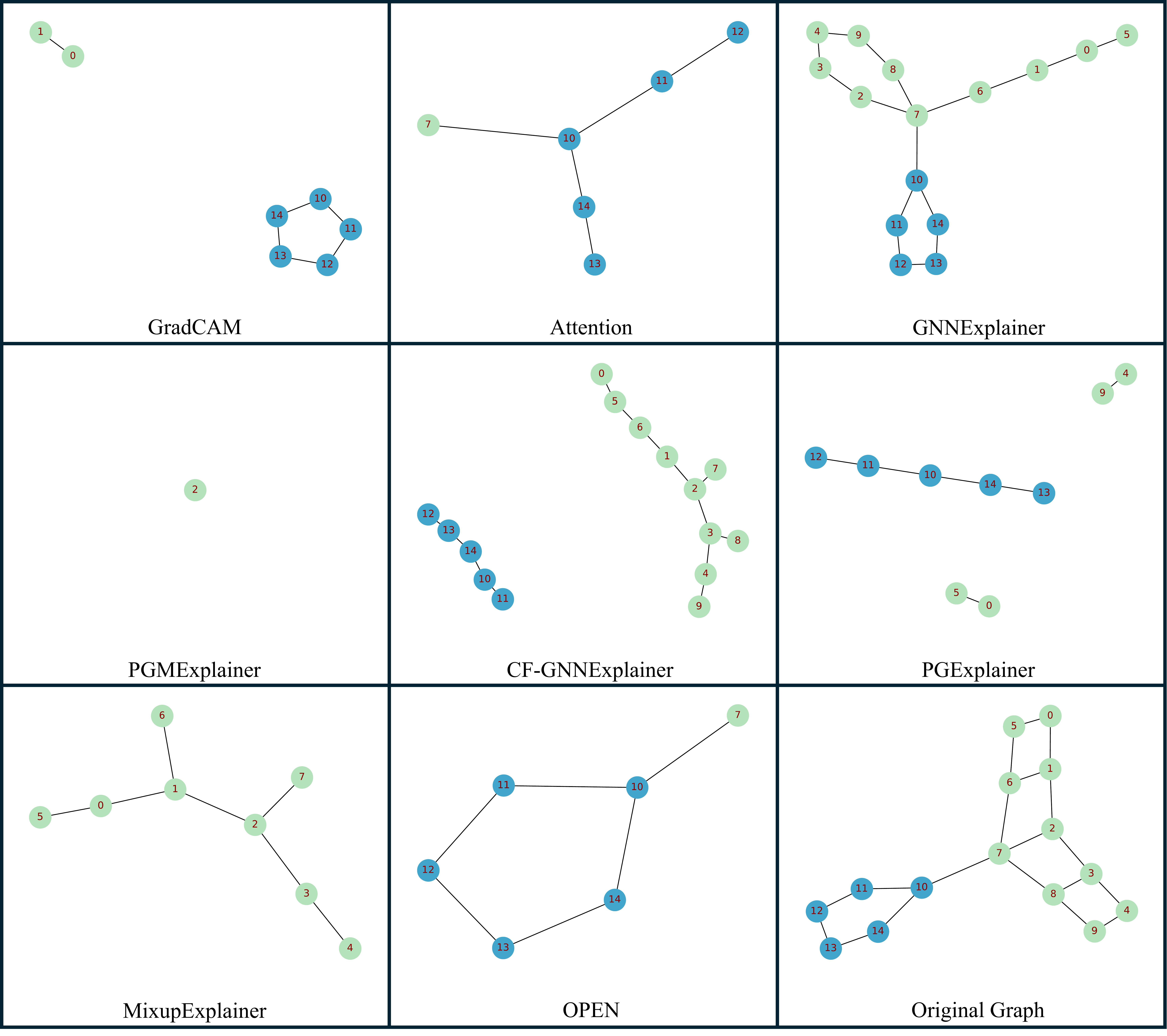}
    \caption{Explanation subgraphs.}
    \label{fig:explanation_subgraph}
\end{figure*}

Figure~\ref{fig:explanation_subgraph} displays explanation subgraphs generated by the proposed \Model~and some other XGNN methods on the Motif dataset in the basis domain. The figure reveals that all methods, except PGMExplainer, successfully identify nodes relevant to predictions (blue nodes). \Model~stands out by considering both node and edge existence probabilities during subgraph generation, ensuring the connectivity of the explanation subgraphs. Moreover, \Model's ability to adjust the size and density of the subgraphs based on prior knowledge offers a more flexible and user-friendly explanation approach compared to existing XGNN methods, enhancing user comprehension.

\end{document}